\documentclass{article}

\PassOptionsToPackage{numbers, compress}{natbib}
\usepackage[preprint]{neurips_2026}

\usepackage[utf8]{inputenc} % allow utf-8 input
\usepackage[T1]{fontenc}    % use 8-bit T1 fonts
\usepackage{hyperref}       % hyperlinks
\usepackage{url}            % simple URL typesetting
\usepackage{booktabs}       % professional-quality tables
\usepackage{amsfonts}       % blackboard math symbols
\usepackage{nicefrac}       % compact symbols for 1/2, etc.
\usepackage{microtype}      % microtypography
\usepackage{xcolor}         % colors

\definecolor{cmlabPrefix}{HTML}{0D207F}
\definecolor{cmlabNumber}{HTML}{0D207F}

\hypersetup{
  colorlinks=true,
  linkcolor=cmlabNumber,
  citecolor=cmlabNumber,
  urlcolor=cmlabNumber,
  pdfborder={0 0 0},
}

\usepackage[capitalize,noabbrev]{cleveref}

\newcommand{\cmlabRef}[4]{%
  {\color{cmlabPrefix}#1}~#3{\color{cmlabNumber}#2}#4%
}
\newcommand{\cmlabRefParen}[4]{%
  {\color{cmlabPrefix}#1}~#3{\color{cmlabNumber}(#2)}#4%
}

\crefformat{figure}    {\cmlabRef{Fig.}{#1}{#2}{#3}}
\Crefformat{figure}    {\cmlabRef{Figure}{#1}{#2}{#3}}
\crefformat{table}     {\cmlabRef{Tab.}{#1}{#2}{#3}}
\Crefformat{table}     {\cmlabRef{Table}{#1}{#2}{#3}}
\crefformat{section}   {\cmlabRef{Sec.}{#1}{#2}{#3}}
\Crefformat{section}   {\cmlabRef{Section}{#1}{#2}{#3}}
\crefformat{subsection}{\cmlabRef{Sec.}{#1}{#2}{#3}}
\Crefformat{subsection}{\cmlabRef{Section}{#1}{#2}{#3}}
\crefformat{equation}  {\cmlabRefParen{Eq.}{#1}{#2}{#3}}
\Crefformat{equation}  {\cmlabRefParen{Equation}{#1}{#2}{#3}}

\usepackage{graphicx}         % figures 
\usepackage{wrapfig,lipsum}

\usepackage{booktabs}
\usepackage{makecell}
\usepackage[table]{xcolor}
\usepackage{amsmath}
\usepackage{multirow}
\usepackage{xcolor}
\usepackage{arydshln}
\usepackage{titletoc}

\usepackage{cleveref}

\crefname{figure}{Fig.}{Figs.}
\Crefname{figure}{Fig.}{Figs.}

\definecolor{Pose}{HTML}{4DB6AC}    % teal — pose anchoring
\definecolor{Domain}{HTML}{FFB74D}  % amber — identity/domain
\definecolor{Dist}{HTML}{BA68C8}    % purple — distribution match

\usepackage{makecell}
\newcommand{\cmark}{\textcolor{green!60!black}{\ding{51}}}
\newcommand{\xmark}{\textcolor{red!80!black}{\ding{55}}}
\usepackage{pifont}                % for \ding

\usepackage{nicematrix}

\definecolor{purple}{RGB}{150, 0, 210} 

\definecolor{myred}{RGB}{132, 10, 50} 

\definecolor{my}{RGB}{220, 53, 98} 

\definecolor{dh}{RGB}{250, 13, 10}

\definecolor{dah}{RGB}{0, 0, 0}
\newcommand{\dah}[1]{\textcolor{dah}{#1}}

\definecolor{bluejh}{RGB}{0, 0, 0}
\newcommand{\jh}[1]{\textcolor{bluejh}{#1}}

\definecolor{greenjh}{RGB}{0, 0, 0}

\definecolor{royalbluejh}{RGB}{0, 0, 0}
\newcommand{\jhfix}[1]{\textcolor{royalbluejh}{#1}}

\definecolor{royalbluejhxx}{RGB}{0, 0, 0}
\newcommand{\jhxx}[1]{\textcolor{royalbluejhxx}{#1}}

\definecolor{royalbluejhxxx}{RGB}{0, 0, 0}

\definecolor{royalbluejhx}{RGB}{0, 0, 0}
\newcommand{\jhx}[1]{\textcolor{royalbluejhx}{#1}}

\definecolor{royalbluejhy}{RGB}{0, 0, 0}

\definecolor{royalbluejhz}{RGB}{0, 0, 0}
\newcommand{\jhz}[1]{\textcolor{royalbluejhz}{#1}}

\definecolor{jhgreenz}{RGB}{0,0,0}
\newcommand{\jhzz}[1]{\textcolor{jhgreenz}{#1}}

\newcommand{\dhmodel}{DipRefGC}

\newcommand{\jhmodel}{RefGC-SR$^2$}

\makeatletter
\newcommand{\sidecaptiontable}[2]{%
  \begingroup
  \def\@captype{table}%
  \caption{#1}%
  \label{#2}%
  \endgroup
}

\newcommand{\sidecaptionfigure}[2]{%
  \begingroup
  \def\@captype{figure}%
  \caption{#1}%
  \label{#2}%
  \endgroup
}
\makeatother

\title{RefGC-SR$^2$: Reference-guided Super-Resolution and\\Refinement of AI Generated Content}

\author{Anonymous Author(s) \\ Affiliation \\ Address \\ \texttt{email}}

\cmlabAuthors{Jeahun Sung$^{1,*}$ \qquad Dahyeon Kye$^{1,*}$ \qquad Soo Ye Kim$^{2,\dagger}$ \qquad Jihyong Oh$^{1,\dagger}$}
\cmlabAffiliations{$^{1}$ CMLab, Chung-Ang University \qquad $^{2}$ Adobe Research}
\cmlabAuthorEmail{\{jhseong,rpekgus,jihyongoh\}@cau.ac.kr \qquad sooyek@adobe.com}
\cmlabProjectPage{https://cmlab-korea.github.io/RefGC-SR2/}

\begin{document}

\maketitle

{
  \renewcommand{\thefootnote}{\fnsymbol{footnote}}
  \footnotetext[1]{Co-first authors (equal contribution).}
  \footnotetext[2]{Co-Corresponding authors.}
}

\begin{figure*}[h!]
	\centering
	% \vspace{-9mm}
	\includegraphics[width=0.85\linewidth]{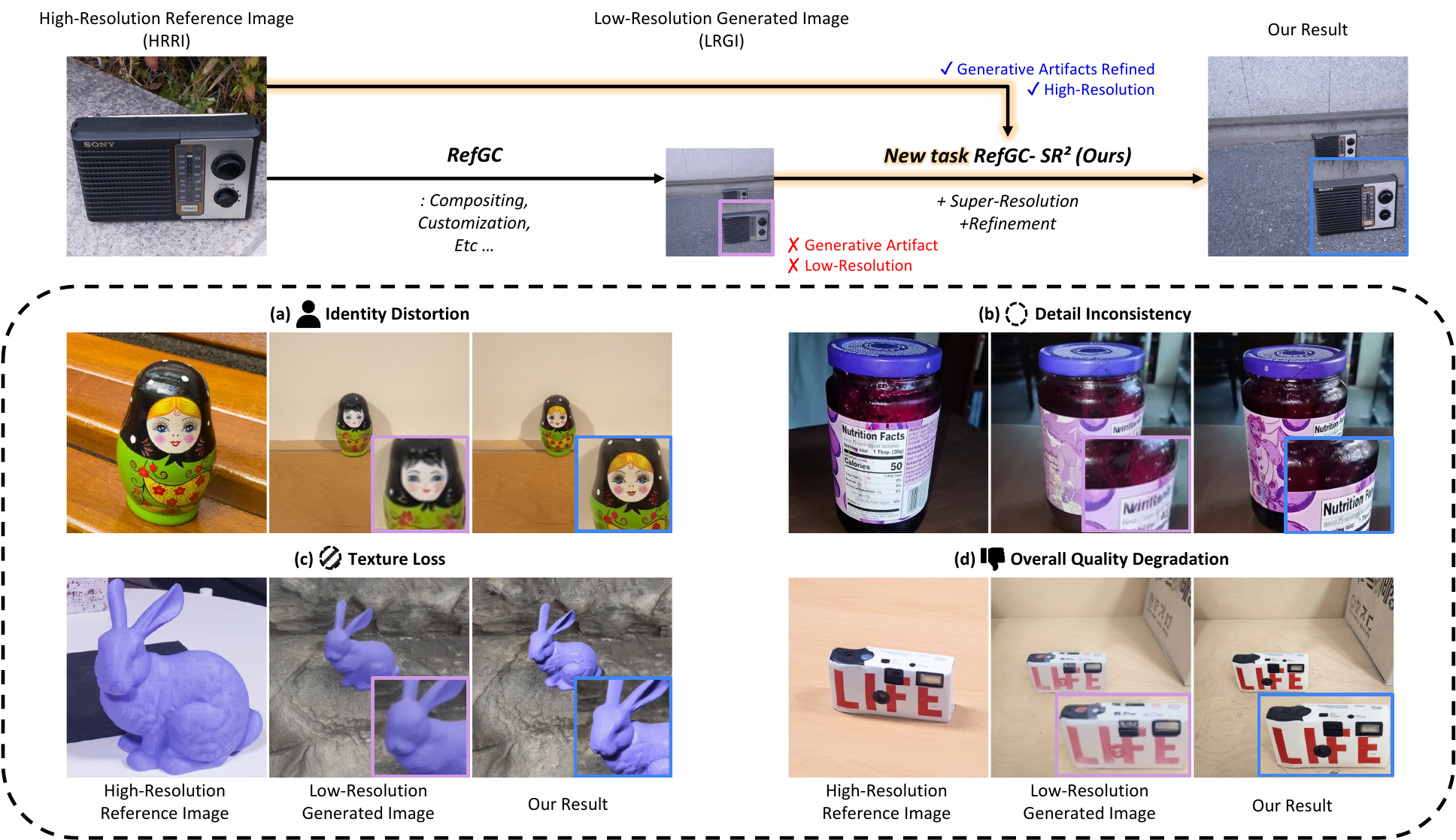}
	% \vspace{-2mm}
    \caption{Reference-guided Generated Content Super-Resolution-Refinement (RefGC-SR$^2$).
    Existing reference-guided generation pipelines first produce a low-resolution generated image (LRGI) from a high-resolution reference image (HRRI), but the output often suffers from four common generative artifacts: \emph{identity distortion}, \emph{detail inconsistency}, \emph{texture loss}, and \emph{overall quality degradation}. We introduce RefGC-SR$^2$, a post-processing task that takes the LRGI and the HRRI as inputs, and produces a high-resolution, artifact-refined output. \textit{Best viewed in zoom.}}
    \label{fig:teaser}
\end{figure*}

\begin{abstract}
    Reference-guided generation (e.g., object compositing, customization) has progressed rapidly, yet current pipelines share a fundamental limitation: the object-centric high-resolution reference image (HRRI) provided by users is downsampled to a fixed low-resolution (LR) before being fed into the model, so the fine-grained details are discarded before the output is even produced. In addition, the generation step then introduces its own artifacts (e.g., identity distortion) on top of this loss. Existing reference-guided generated content refinement (RefGCR) methods can correct some of these artifacts but still operate in the LR domain; reference-guided super-resolution (RefSR) methods recover resolution but assume natural-image degradations and ignore the artifact distribution of generative pipelines. To address both gaps in a single formulation, we introduce a new task: \emph{reference-guided generated content super-resolution-refinement} (RefGC-SR$^2$), where the original HRRI is reused at the post-processing stage to recover lost details, refine generative artifacts, and upscale the output simultaneously. We construct the first real-world triplet data generation pipeline for this RefGC-SR$^2$ task, training a diptych-conditioned generator to synthesize paired low-quality anchors that public pretrained models cannot provide. We further present a frequency-aware diffusion transformer model for \jhmodel{} that selectively injects fine details from the HRRI while removing generative artifacts. Extensive experiments demonstrate that our \jhfix{\jhmodel{} model} successfully (i) refines the object identity faithfully \jhx{with respect} to the reference, and (ii) recovers high-resolution details, so that the final result is significantly higher quality and practically more usable compared to existing RefGCR and RefSR baselines.
\end{abstract}
\section{Introduction}
\label{sec:intro}

Image generation has rapidly evolved into a user-conditioned process, where users provide not only text prompts but also masks, structural controls, and reference images to personalize and control the output~\citep{brooks2023instructpix2pix, zhang2023adding, ye2023ip}. In particular, reference-guided generation has become a practical paradigm for user-driven visual creation, supporting applications such as image editing~\citep{liu2025magicquill, yang2023paint, chen2024zero}, customization~\citep{feng2026personalize, zhang2025freecus, tarres2025multitwine, tan2025ominicontrol, wu2025less}, and compositing~\citep{huang2025dreamfuse, chen2024anydoor, song2026insert, li2025aicomposer, wang2025unicombine, song2023objectstitch}. However, existing reference-guided generated content (RefGC) pipelines remain fundamentally limited in leveraging high-resolution reference images (HRRIs), which significantly limits practical adoption. \jhx{Specifically}, in real-world scenarios, users often possess HRRIs that contain rich visual information. \jhx{ However,} most RefGC pipelines downsample these references to a fixed low-resolution (LR) before injecting them into the generation process, e.g., $224^2$ visual tokens~\citep{ye2023ip, chen2024anydoor} or $512^2$ diffusion inputs~\citep{tan2025ominicontrol}. As a result, high-frequency (HF) information in the HRRI is largely discarded before generation begins. Then as shown in Fig.~\ref{fig:teaser}, the generated content (GC) further suffers from recurring generative artifacts: \textit{identity distortion}, \textit{detail inconsistency}, \textit{texture loss}, and \textit{overall quality degradation}. In other words, although users provide rich HRRIs, current RefGC pipelines fail to fully transfer their fine-grained information to the final GC output.

\paragraph{Positioning of Our Work} To position our work in the right context, we first categorize a wide range of related image enhancement tasks in Table~\ref{tab:task_comparison}. Image Super-Resolution (ISR)~\citep{choi2025framer, duan2025dit4sr, dong2025tsd} recovers high-resolution (HR) images from LR inputs, but assumes natural-image degradations such as bicubic downsampling, blur, noise, compression, or realistic camera degradations. Reference-guided SR (RefSR)~\citep{lee2025reference, lee2024refqsr} further exploits an external HRRI, but is still formulated for natural-image SR rather than GC with generative artifacts. Generated content super-resolution (GCSR)~\citep{jeong2025latent, du2024demofusion, tragakis2024one} aims to overcome the native resolution limits of generative models, but operates on the GC alone without any HRRI. Reference-guided generated content SR (RefGCSR), which is yet to be explored in literature, can further use an HRRI for GCSR, but it is not designed to jointly refine RefGC-specific generative artifacts. On the other hand, image restoration (IR)~\citep{ye2024learning}, generated content restoration (GCR)~\citep{lin2024diffbir}, and reference-guided refinement (RefR)~\citep{zhang2024reference, guo2024refir} refine degraded or artifact-contaminated images, but do not jointly perform SR with an HRRI for LR RefGC inputs. 
Most closely related, reference-guided generated content refinement (RefGCR)~\citep{song2024refine, ouyang2025consistency, zhou2026refineanything, liu2025omnirefiner} uses a reference image to correct generative artifacts in GC, but \jhzz{it is} primarily a fixed-resolution refinement task and \jhzz{does} not recover an HR output from an LR GC input. Detailed discussions of these related tasks are provided in Appendix~\ref{app:related_work}.

As shown in Table~\ref{tab:task_comparison}, existing tasks satisfy only a subset of the four criteria required by our setting: a GC image and an HRRI as inputs, and SR and generative artifact refinement as joint target operations. In contrast, our newly defined \emph{Reference-guided Generated Content Super-Resolution-Refinement} (RefGC-SR$^2$) is the first task that jointly satisfies all four criteria, making it a distinct post-processing task well-suited for contemporary RefGC pipelines.
\begin{table*}[t]
\centering
\caption{Comparison of related image enhancement tasks. RefGC-SR$^2$ is the first task that jointly satisfies all four criteria: a generated content (GC) input and an HR reference image (HRRI) input, performing super-resolution (SR) and artifact refinement simultaneously.
}
\label{tab:task_comparison}
\scriptsize
\setlength{\tabcolsep}{5.2pt}
\renewcommand{\arraystretch}{1.18}
\begin{NiceTabular}{lccccccccw{c}{1.5cm}}
\CodeBefore
    \columncolor{Domain!10}{10}
\Body
\toprule
\multirow{2}{*}{Task}
& \multicolumn{4}{c}{Super-Resolution Category}
& \multicolumn{4}{c}{Refinement Category}
& \multirow{2}{*}{\makecell{\textbf{RefGC-SR$^2$}\\\textbf{(Ours)}}} \\
\cmidrule(lr){2-5} \cmidrule(lr){6-9}
& $\mathrm{I\textbf{SR}}$ 
& $\mathrm{Ref\textbf{SR}}$ 
& $\mathrm{GC\textbf{SR}}$ 
& $\mathrm{RefGC\textbf{SR}}$ 
& $\mathrm{I\textbf{R}}$
& $\mathrm{Ref\textbf{R}}$ 
& $\mathrm{GC\textbf{R}}$ 
& $\mathrm{RefGC\textbf{R}}$ 
& \\
\midrule
GC
& \xmark & \xmark & \cmark & \cmark
& \xmark & \xmark & \cmark & \cmark
& \cmark \\
HRRI
& \xmark & \cmark & \xmark & \cmark
& \xmark & \cmark & \xmark & \cmark
& \cmark \\
SR
& \cmark & \cmark & \cmark & \cmark
& \xmark & \xmark & \xmark & \xmark
& \cmark \\
Refinement
& \xmark & \xmark & \xmark & \xmark
& \cmark & \cmark & \cmark & \cmark
& \cmark \\
\bottomrule
\end{NiceTabular}

\vspace{-0.35cm}
\end{table*}
RefGC-SR$^2$ takes (i)~a low-resolution generated content image (LRGI) from a RefGC pipeline and (ii)~the original HRRI, and generates an HR image that preserves the HRRI's fine-grained information while refining generative artifacts in the LRGI. 
It targets a practical yet underexplored stage of modern reference-guided generation pipelines, converting an initial GC output into an HR, reference-faithful final image that preserves object identity in HRRI and recovers fine-grained details, helping to achieve the last mile of personalized image editing for practical adoption.

A challenge in training a RefGC-SR$^2$ model is acquiring paired triplets. Note that such datasets do not exist in standard SR or refinement datasets~\citep{liu2025omnirefiner, zhou2026refineanything}. In each triplet, HRGT denotes the desired high-resolution ground-truth target, HRRI provides reference-specific details of the same object instance (corresponds to object-centric user-provided reference image), and LRGI serves as the LR input containing generative artifacts. This construction is non-trivial: existing SR datasets use hand-crafted degradations rather than real RefGC artifacts, while existing GC datasets usually lack paired HRGTs and HRRIs. 
Moreover, naively using off-the-shelf RefGC models~\citep{huang2025dreamfuse, zhang2025freecus, feng2026personalize, song2026insert} to synthesize LRGIs often produces outputs whose object pose differs from HRGT. Such pose mismatch is unsuitable for our post-processing task as users simply wish to improve the quality and resolution of the generated result \textit{as is} without \textit{changing the pose} of the generated object. Thus, the object pose should be aligned between LRGI and HRGT, allowing the model to focus on artifact refinement and resolution recovery rather than learning an unintended pose correction task.

To this end, we construct RefGC-SR$^2$ Dataset, the first real-world LRGI-HRRI-HRGT triplet dataset tailored to this task. The HRRI and HRGT depict the same object instance under different views, poses, or scene contexts, and are collected from real-world datasets~\citep{kim2025orida, liu2025uncommon, xue2025ultravideo} and curated with a vision-language model (VLM)~\citep{bai2025qwen3}. 
To synthesize an aligned LRGI for each HRRI-HRGT pair, we introduce DipRefGC (Diptych-Conditioned RefGC Generator). DipRefGC generates an LRGI that inherits the object appearance from HRRI while matching the object pose of HRGT, thereby producing a realistic artifact-containing RefGC output for supervised training.

% \jhtwo{Based on the constructed RefGC-SR$^2$ Dataset, 
Furthermore, we propose the \textit{first} RefGC-SR$^2$-targeted model that is trained on RefGC-SR$^2$ Dataset. 
Our \jhfix{\jhmodel{} Model} consists of two main components: (i) frequency-adaptive mixture of LoRA experts (FreqMoLE) and (ii) frequency-\jhzz{adaptive} loss ($\mathcal{L}_f$).
% FreMoLE is motivated by the layer-wise frequency hierarchy of FLUX-Kontext~\citep{labs2025flux}, where early DiT layers capture global structure and later layers refine fine details (Fig.~\ref{fig:freq_motiv}-(a)). 
FreqMoLE is motivated by the layer-wise frequency hierarchy of FLUX-Kontext~\citep{labs2025flux}, where early DiT blocks capture global structure and later blocks refine fine details. 
It introduces two LoRA experts: a low-frequency (LF) expert and \jhzz{an} HF expert. 
A routing gate adaptively controls their weights, assigning higher weight to the LF expert in early layers and to the HF expert in later layers.
Our $\mathcal{L}_f$ is designed based on the frequency relationship among LRGI, HRRI, and HRGT. 
It guides the model to learn global structure aligned with HRGT during training, while enabling the transfer of fine details from HRRI at inference. Our contributions can be summarized as follows:
\begin{itemize}
    \item We formulate RefGC-SR$^2$ problem, the first post-processing task that reuses the user-provided object-centric HRRI as a recovery source to jointly upscale and refine RefGCs.
    \item We construct RefGC-SR$^2$ Dataset, the first real-world LRGI-HRRI-HRGT triplet dataset supporting supervised training and evaluation of \jhzz{the} RefGC-SR$^2$ task.
    \item We propose the \textit{first} model for RefGC-SR$^2$ with frequency-aware modules \jhzz{inserted into} DiT blocks to effectively remove artifacts in LRGI while increasing resolution.
\end{itemize}

% To mitigate the artifacts above, recent reference-guided \textit{refinement} methods~\citep{song2024refine, ouyang2025consistency} post-process the generated output using the reference image. While they partially recover details and suppress some artifacts, two limitations remain. First, they tend to over-rely on the reference image, leaving \emph{detail loss} and \emph{identity distortion} only partially mitigated. Second, they operate at the low-resolution (LR) and therefore cannot resolve \emph{hallucinated structures}, \emph{boundary artifacts}, and \emph{texture wash-out} while restoring HR detail. Conversely, classical reference-based super-resolution (SR) methods~\citep{jeong2025latent, du2024demofusion, tragakis2024one} recover resolution but assume natural-image degradations (bicubic, blur, JPEG) and are blind to the artifact distribution of 
% reference-guided generation pipelines.
\section{RefGC-SR$^2$ Dataset}
\label{sec:dataset}
% users simply wish to improve the quality and resolution of the generated result as is without changin
% Training a RefGC-SR$^2$ model requires triplets that faithfully reflect real-world reference-guided generation scenarios.
% Moreover, users provide an object-centric HRRI and expect the GC to be improved in quality and resolution, \textit{without} modifying the object pose. 
% From a training-data perspective, this imposes three key requirements. 
We define three key requirements for constructing RefGC-SR$^2$ data triplets (HRRI, HRGT, LRGI) such that \jhzz{they are} closely aligned to the reference-guided generation and refinement pipeline in real-world scenarios:
First, HRRI and HRGT should depict the \emph{same} object instance in \textit{real-world} images, while \textit{differing} in view, pose, or scene context. Second, the input LRGI should be \jhzz{an} LR GC image containing generative artifacts as shown in Fig.~\ref{fig:teaser}. Third, LRGI should be aligned (with object pose preserved) with the corresponding HRGT, ensuring that the model focuses on SR and refinement tasks rather than unintended pose correction.
Note that no existing datasets satisfy these requirements simultaneously. SR datasets rely on hand-crafted degradations and lack generative artifacts, while recent refinement datasets~\citep{liu2025omnirefiner, zhou2026refineanything} construct corrupted inputs using \jhzz{image editing model}~\citep{wu2025qwen}, which do not capture the true distribution of RefGCs. Off-the-shelf compositing or customization models cannot be reused as data generators, since they alter the object pose. 
As a result, they are insufficient for generating LRGIs that both contain realistic generative artifacts and preserve the object pose of the HRGT, as required by our task.
% As a result, they are insufficient for modeling both the generative artifact and the consistency of object pose between HRRI and HRGT required in our task. 
We construct the dataset in two stages:  (i) real-world HRRI-HRGT pair curation and (ii) LRGI synthesis with our DipRefGC.

\begin{figure*}[t]
    \centering
    \includegraphics[width=0.9\linewidth]{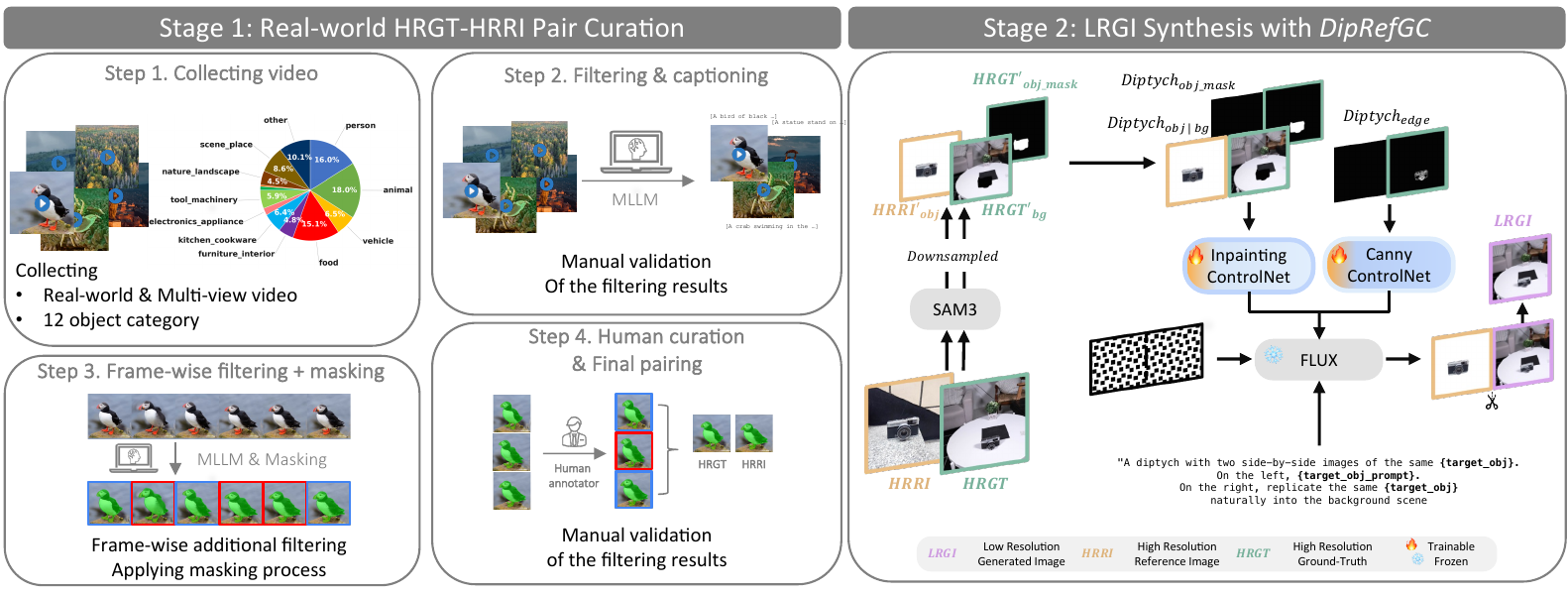}
    % \caption{Construction pipeline of our RefGC-SR$^2$ Dataset. In Stage 1 (Sec.~\ref{sec:hrgt-hrri-curation}), we curate real-world HRRI-HRGT pairs from object-centric multi-view videos through video collection, VLM-based~\citep{bai2025qwen3} filtering and captioning, frame-wise filtering with masking, and final human validation. In Stage 2 (Sec.~\ref{sec:DipRefGC}), we synthesize the corresponding LRGI using our proposed DipRefGC, which takes the HRRI object appearance and HRGT-derived pose condition as inputs.
    % The inpainting ControlNet~\citep{controlnet-inpaint} preserves the reference object appearance, while the Canny ControlNet~\citep{controlnet-canny} enforces the target pose from HRGT.
    % This produces pose-consistent, generative artifact-containing LRGIs paired with HRRI and HRGT, enabling supervised training for RefGC-SR$^2$.}
    \caption{Construction pipeline of our RefGC-SR$^2$ Dataset.
    Stage 1 (Sec.~\ref{sec:hrgt-hrri-curation}) curates real-world HRRI-HRGT pairs from object-centric multi-view videos through VLM-based filtering~\citep{bai2025qwen3}, masking, and human validation. Stage 2 (Sec.~\ref{sec:DipRefGC}) synthesizes the corresponding LRGI with DipRefGC, which combines HRRI appearance with HRGT-derived pose conditions using Inpainting~\citep{controlnet-inpaint} and Canny ControlNets~\citep{controlnet-canny}. This produces pose-consistent, generative artifact-containing LRGIs paired with HRRI and HRGT, enabling supervised training for RefGC-SR$^2$. \textit{Best viewed in zoom.}}
    \label{fig:hrri_hrgt_paring}
    \vspace{-1em}
\end{figure*}

\subsection{Stage 1: Real-world HRRI-HRGT Pair Curation}
\label{sec:hrgt-hrri-curation}

We collect HRRI-HRGT pairs from higher-resolution ($1024^2$ to $2048^2$) real-world image and video datasets that cover diverse object categories, scene contexts, and viewpoints. ORIDa~\citep{kim2025orida} provides real compositing pairs of objects in different backgrounds. 
uCO3D~\citep{liu2025uncommon} offers multi-view captures of objects across everyday categories. 
For the video dataset uCO3D, we apply additional processing to construct HRRI-HRGT pairs, as shown in Fig.~\ref{fig:hrri_hrgt_paring}-Stage 1. We first collect the videos (Step 1). Next, we filter object-centric videos using Qwen3-VL~\citep{bai2025qwen3} (Step 2). We then perform frame-wise filtering with Qwen3-VL and generate object masks using SAM3~\citep{carion2025sam} (Step 3). Finally, human annotators conduct manual verification (Step 4). Through this process, we obtain high-quality HRRI-HRGT pairs for training. \dah{See Appendix~\ref{sec:dataset_statistic} for further analysis.}

% \begin{wrapfigure}{r}{0.35\linewidth}
%     % \vspace{-5mm}
%     \centering
%     \includegraphics[width=0.89\linewidth]{figs/triplet_new.pdf}
%     % \vspace{-3mm}
%     \caption{\small
%     RefGC-SR$^2$ triplet examples with HRRI, HRGT and LRGI. 
%     }
%     \label{fig:diprefgc_triplet}
%     \vspace{-5mm}
% \end{wrapfigure}

\subsection{Stage 2: LRGI Synthesis with DipRefGC Using HRRI-HRGT Pairs}
\label{sec:DipRefGC}

Given an HRRI-HRGT pair, a seemingly straightforward way to obtain an LRGI is to feed the pair into off-the-shelf RefGC models~\citep{feng2026personalize, song2026insert} and downsample the outputs. However, since the goal of our RefGC-SR$^2$ task is to jointly recover resolution and refine generative artifacts of the RefGCs, this naive pipeline is insufficient as the LRGI will unavoidably modify the object pose of the corresponding HRGT. We therefore impose a pose constraint on LRGI synthesis
% LRGI and HRGT should share the same object pose, 
so that the model correctly learns SR and artifact refinement rather than an unintended pose correction. 
Note that existing RefGC models~\citep{huang2025dreamfuse, feng2026personalize, song2026insert, zhang2025freecus, chen2024anydoor} have no explicit control over object pose.
% , as the generated object may deviate from the pose of HRGT.
Thus, we propose DipRefGC (Diptych-Conditioned RefGC Generator), built on a frozen FLUX backbone~\citep{flux1-dev} adapted with LoRA-tuned~\citep{hu2022lora} dual ControlNets~\citep{zhang2023adding}. DipRefGC synthesizes LRGIs that inherit the object appearance from HRRI while following the object pose of HRGT, thereby producing LRGIs with generative artifacts suitable for triplet supervision in RefGC-SR$^2$.
Before DipRefGC synthesis, we downsample the curated HRRI-HRGT pairs to $512^2$ resolution, while retaining the original HR images as HRRI and HRGT in the final triplets. DipRefGC adopts a diptych~\citep{shin2025large} formulation with two specialized ControlNets~\citep{zhang2023adding} to disentangle appearance and pose control. 
\begin{wrapfigure}{r}{0.3\linewidth}
    % \vspace{-5mm}
    \centering
    \includegraphics[width=0.89\linewidth]{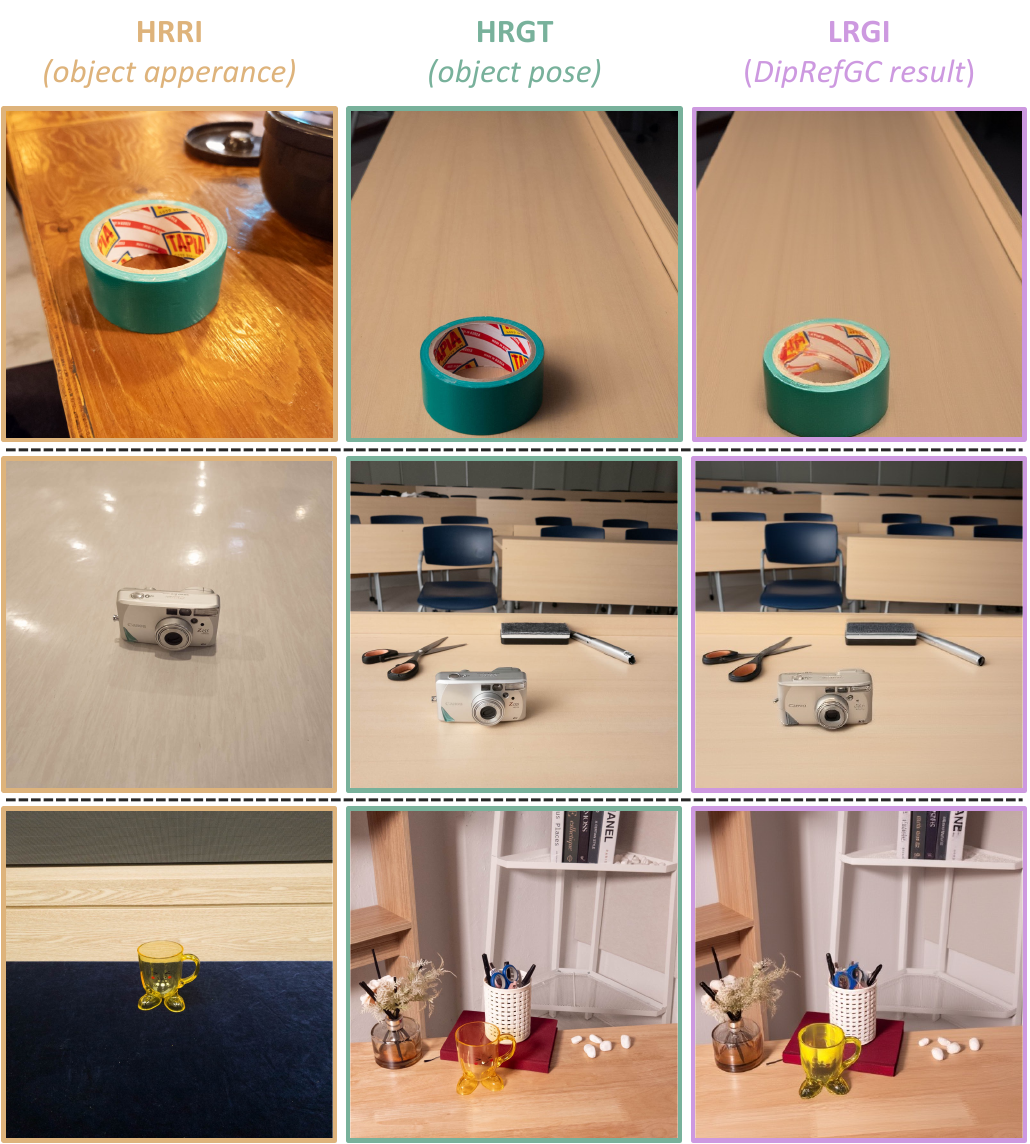}
    % \vspace{-3mm}
    \caption{\small
    RefGC-SR$^2$ triplet examples (HRRI, HRGT and LRGI). 
    }
    \label{fig:diprefgc_triplet}
    \vspace{-5mm}
\end{wrapfigure}
As shown in Fig.~\ref{fig:hrri_hrgt_paring}-Stage 2, we construct two diptych-style conditioning inputs, each composed of a left reference panel and a right generation panel.
For appearance control, the \emph{Inpainting ControlNet}~\citep{controlnet-inpaint} receives a diptych input whose left panel contains the segmented object from HRRI, while the right panel contains the masked HRGT background and a tight object mask.
We mask out the object in \textit{HRGT} so that the model needs to rely on the object in HRRI.
This set-up guides the model to resynthesize the object inside the target region in HRGT based on the object appearance in \textit{HRRI}, effectively mimicking the generative artifacts (e.g., object identity distortion) of RefGC approaches. 
For pose control, the \emph{Canny ControlNet}~\citep{controlnet-canny} receives a second diptych input whose right panel contains the HRGT edge map restricted to the object region. This condition enforces the generated object to be aligned to the HRGT. This separation allows DipRefGC to independently control \emph{what} to generate (appearance) and \emph{how} to generate it (pose and structure).
We fine-tune DipRefGC on the curated HRRI-HRGT pairs using the standard flow-matching objective~\citep{lipman2022flow} of the FLUX backbone, where the reconstruction loss is applied only to the object region.
We then run the trained DipRefGC on the curated HRRI-HRGT pairs to construct the final RefGC-SR$^2$ Dataset, consisting of 40K training triplets and \jhzz{a} 200-sample evaluation split. As shown in Fig.~\ref{fig:diprefgc_triplet}, each triplet provides an object pose-consistent, generative artifact-\jhzz{containing} $512^2$ LRGI, its corresponding HRRI, and the target HRGT, enabling supervised training of RefGC-SR$^2$ model.

\section{\jhmodel{} Model}
\label{sec:method}

\begin{figure*}[t!]
	\centering
	\includegraphics[width=0.9\linewidth]{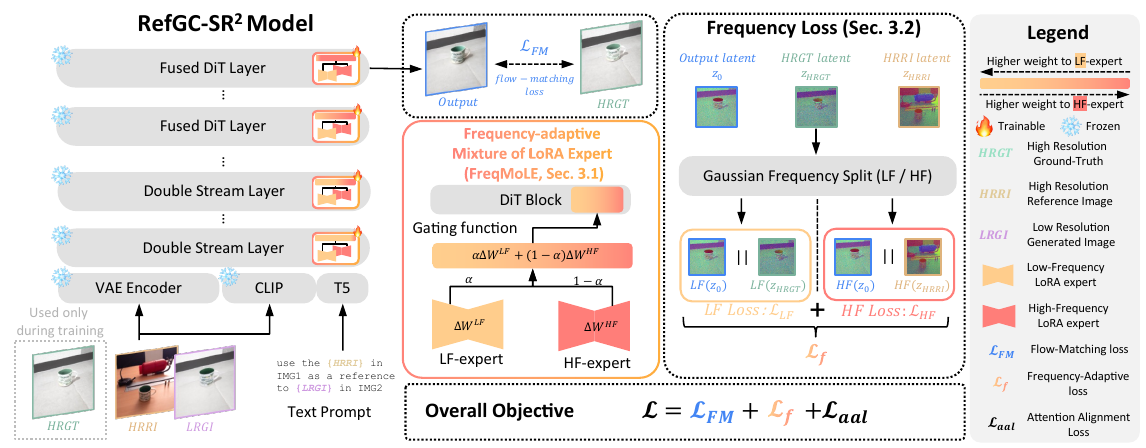}
	\caption{Overview of our \jhfix{\jhmodel{} model}. The frozen VAE encodes the inputs (LRGI, HRRI, HRGT (train only)) and T5 encodes the text prompt. We insert trainable FreqMoLE modules into a frozen FLUX-Kontext~\citep{labs2025flux} backbone, supervised by our \jhzz{frequency-adaptive} loss $\mathcal{L}_f$.
    }
    \label{fig:main_model}
    \vspace{-5mm}
\end{figure*}

We propose \jhzz{the} \jhmodel{} Model addressing the \jhmodel{} task, which jointly removes generative artifacts and increases resolution.
RefGC-SR$^2$ is particularly challenging because the model must not only recover HR details from an LRGI, but also refine generative artifacts while selectively exploiting a clean reference image (HRRI) captured under different conditions. 
Our method consists of two key components: 
(1) a \emph{frequency-adaptive Mixture-of-LoRA Experts} (FreqMoLE) applied to all DiT blocks, and 
(2) a \emph{frequency-\jhzz{adaptive} loss} ($\mathcal{L}_f$) that supervises the experts with band-specific objectives. 
The overall pipeline is illustrated in Fig.~\ref{fig:main_model}.

\begin{figure*}[h!]
	\centering
	\includegraphics[width=0.85\linewidth]{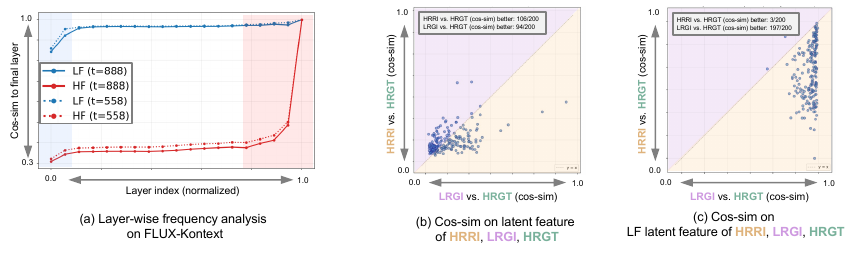}
	\caption{\jh{Motivation experiments for our method: 
(a) shows the layer-wise frequency characteristics of the FLUX-Kontext model~\citep{labs2025flux}. 
(b) compares the similarity between latent features of LRGI, HRRI, and HRGT. 
(c) shows the low-frequency similarity between their latent features.}}
    \label{fig:freq_motiv}
    \vspace{-1.0em}
\end{figure*}

\paragraph{Motivations}\label{method:motivations}
\jhx{We design our \jhfix{\jhmodel{} model} based on two key observations from FLUX-Kontext~\citep{labs2025flux} and our RefGC-SR$^2$ Dataset.}
% We design our \jhfix{\jhmodel{} model} based on two key observations from FLUX-Kontext~\citep{labs2025flux} and reference-guided super-resolution data (in our RefGC-SR$^2$ dataset). 
% \jhfix{
% These observations are conducted to address two concerns before designing the \jhmodel{} model. 
% The first concern is the unknown 
First, to better understand the behavior of the backbone model (FLUX-Kontext) when applied to the \jhmodel{} task, we analyze the frequency characteristics of our backbone.  
% The second is how to effectively exploit the relationship among HRGT, HRRI, and LRGI.
% To address the first concern, we analyze the frequency characteristics of FLUX-Kontext. 
% We measure LF and HF energy from latent activations of diffusion transformer layers. 
\jhzz{We measure the cosine similarity between the LF/HF components of intermediate-layer latent activations and their counterparts in the final layer.}
As shown in Fig.~\ref{fig:freq_motiv}-(a), 
LF quickly saturates within the first $\sim$5\% of layers, while HF emerges sharply in the last $\sim$10\%. 
This indicates that global structure (LF) is formed in early layers, whereas fine details (HF) are formed in later layers. 
This observation motivates our design of FreqMoLE (Sec.~\ref{subsec:freqmole}), which combines two band-specific LoRA experts, $\Delta W^{LF}$ and $\Delta W^{HF}$, using a coarse-to-fine, \jhx{layer-}depth-dependent gating $\alpha$.
Second, to effectively exploit the relationship among LRGI, HRRI and HRGT in our model design,
% To address the second concern, 
we analyze their relation in the latent space. 
We compute cosine similarity to measure which input is closer to HRGT in Fig.~\ref{fig:freq_motiv}-(b), where the results show no clear trend across all features. 
However, as shown in Fig.~\ref{fig:freq_motiv}-(c), when applying a frequency cut-off to isolate LF components, LRGI features are consistently closer to HRGT than HRRI.
This observation motivates the use of frequency-aware supervision, leading to our {frequency-adaptive} loss ($\mathcal{L}_f$, Sec.~\ref{sec:method_loss}).
% We design \jhfix{\jhmodel{} model} based on two observations from FLUX-Kontext~\citep{labs2025flux} and reference-guided super-resolution data. 
% The first concerns where different frequency components are formed across network layers. 
% The second examines the frequency similarity among HRGT, HRRI, and LRGI.

% \jh{We measure low-frequency (LF) and high-frequency (HF) energy from latent activations at each transformer block (Fig.~\ref{fig:freq_motiv}-(a)). 
% LF quickly saturates within the first $\sim$5\% of layers, while HF emerges sharply in the last $\sim$10\%. 
% The gap between the two exceeds 0.55 in the middle layers. 
% This indicates that LF (global structure) is formed in early layers, whereas HF (fine details) is formed in later layers. 
% This observation motivates our FreqMoLE, which combines two band-specific LoRA experts ($\Delta W^{LF}$, $\Delta W^{HF}$) using depth-dependent gating $\alpha$ in a coarse-to-fine manner.}

% \jh{We further analyze which input is closer to HRGT using cosine similarity in the latent space (Fig.~\ref{fig:freq_motiv}-(b)). 
% The results do not show a clear trend across features. 
% However, as shown in Fig.~\ref{fig:freq_motiv}-(c), when applying a frequency cut-off to isolate LF components, LRGI features are consistently closer to HRGT than HRRI. 
% This observation motivates the use of frequency-aware supervision, leading to our frequency-based loss ($\mathcal{L}_f$-~Sec.~\ref{sec:method_loss}).}

\paragraph{Overall Pipeline}\label{sec:method_pipeline} 
\jhmodel{} model takes three inputs: an LRGI of the target viewpoint, an HRRI of the same instance from a different viewpoint, and \jhfix{a text instruction that specifies the task of refining and super-resolving the LRGI using the HRRI}.
Our goal is to refine and upscale (4$\times$) the LRGI with the guidance of HRRI, so that the result is close to HRGT.
As shown in Fig.~\ref{fig:main_model}, LRGI and HRRI are encoded using a frozen VAE~\citep{kingma2013auto}, and the prompt encoder follows ImageCritic~\citep{ouyang2025consistency}. To exploit the layer-wise frequency hierarchy of the FLUX-Kontext~\citep{labs2025flux} model described above, we freeze the backbone and apply \jh{our proposed} Frequency-adaptive Mixture of LoRA Experts (FreqMoLE) to all MMDiT layers. FreqMoLE injects frequency information in a layer-dependent manner, emphasizing LF components in early layers and HF components in later layers. Furthermore, we introduce a {frequency-adaptive} loss ($L_f$) as a guidance term to effectively utilize information from both HRRI and HRGT. It uses LF information from HRGT to guide global structure, and HF information from HRRI to restore fine details.

\subsection{Frequency-adaptive Mixture of LoRA Experts (FreqMoLE)}
\label{subsec:freqmole}

Based on the observation on layer-wise frequency hierarchy in Sec.~\ref{method:motivations}-\textbf{Motivations}, we replace a single LoRA in each DiT block with two \textit{band-specialized} experts: \jhzz{an} LF expert $\Delta W^{LF}$ and a HF expert $\Delta W^{HF}$. 
The outputs of the two experts are combined using a depth-dependent gating function $\alpha$: $W^{FreqMoLE}=\alpha \,\Delta W^{LF} + (1 - \alpha)\,\Delta W^{HF}$.
The gating value $\alpha$ is parameterized as a learnable scalar for each layer and initialized with a coarse-to-fine prior. 
It is set close to $1.0$ in earlier layers to favor LF components and approaches $0$ in later layers to emphasize HF components (Sec.~\ref{sec:method}-\textbf{Motivations}). 
\jhx{The gating function is frozen in the early stage of training to keep LF dominance in early layers and HF dominance in later layers, and is then jointly optimized with the experts.}
% \jhx{The gating function} is frozen at an earlier stage of training to encourage specialization, and then jointly optimized with the experts. Through this design, FreqMoLE adaptively injects frequency-aware information in a coarse-to-fine manner, aligned with the layer-wise frequency hierarchy.

\subsection{frequency-adaptive Loss ($\mathcal{L}_f$)}
\label{sec:method_loss}

To encourage band-specific behavior of the experts, we introduce a {frequency-adaptive} loss ($\mathcal{L}_f$) applied at the \textit{latent} level. 
The goal of this loss is to promote frequency decomposition between the LF and HF experts, while guiding the LRGI input to learn LF global structure from the HRGT and HF details from the HRRI. We first decompose each latent $z$ into LF and HF components using Gaussian band split~\citep{burt1983laplacian}: $\mathrm{LF}(z) \;=\; G_\sigma * z, \text{and  } \mathrm{HF}(z) \;=\; z - \mathrm{LF}(z)$. The LF term aligns the global structure with the HRGT: $\mathcal{L}_{LF} \;=\; \big\| \mathrm{LF}(z_0) - \mathrm{LF}(z_{HRGT}) \big\|_{\mathcal{M}},$ where $\|\cdot\|_{\mathcal{M}}$ denotes the $\ell_1$ norm restricted by an object mask. The HF term transfers fine detail from the HRRI. 
Since the HRRI and HRGT differ in viewpoint, we avoid pixel-wise alignment and instead match channel-wise statistics~\citep{li2017demystifying}:
$\mathcal{L}_{HF} \;=\; \Big| \mu_{\mathcal{M}}\!\big(\mathrm{HF}(z_0)\big) 
                       - \mu_{\mathcal{M}}\!\big(\mathrm{HF}(z_{HRRI})\big) \Big|
                  \;+\; \Big| \sigma_{\mathcal{M}}\!\big(\mathrm{HF}(z_0)\big) 
                       - \sigma_{\mathcal{M}}\!\big(\mathrm{HF}(z_{HRRI})\big) \Big|.$
The \jhfix{frequency-adaptive loss ($\mathcal{L}_f$)} is defined as:
$\mathcal{L}_{f} \;=\; \lambda_{LF}\,\mathcal{L}_{LF} \,+\, \lambda_{HF}\,\mathcal{L}_{HF}.$
$\mathcal{L}_{LF}$ mainly updates the LF expert through the $\alpha$-weighted path, while $\mathcal{L}_{HF}$ mainly updates the HF expert through the $(1-\alpha)$ path. 
Combined with the layer-wise prior, this enables smooth specialization without hard routing. For the full loss (objective function) including this term, see the \jhz{Appendix~\ref{sec:aux_losses}} and Overall Objective in Fig.~\ref{fig:main_model}.

\section{Experiments}
\label{sec:experiment}

% \subsection{Experimental Set-up}
\begin{table}[t]
\centering
\scriptsize
\caption{Quantitative comparison on RefGC-SR$^2$. 
The best and second-best results are highlighted in \textbf{bold} and \underline{underline}, respectively. The Reference column indicates whether the model uses an additional HRRI. Asterisk (*) denotes that the model has been finetuned on our dataset.}
\label{tab:main_quan}
\begin{tabular}{clcccccc}
\toprule
\textbf{Task}
& \textbf{Model} 
& \textbf{Reference}
& \textbf{CLIP-I$\uparrow$} 
& \textbf{DINO$\uparrow$} 
& \textbf{PSNR$\uparrow$} 
& \textbf{SSIM$\uparrow$} 
& \textbf{LPIPS$\downarrow$} \\
\midrule
\multirow{4}{*}{SR}
& DiT4SR (ICCV'25) 
& \xmark
& 0.8156
& 0.6555
& 15.3762
& 0.4932
& 0.4282 \\

& DiT4SR$^{*}$ (ICCV'25) 
& \xmark
& 0.8186 
& 0.6545 
& 15.1005 
& 0.5726 
& 0.3884 \\

& TSD-SR (CVPR'25)
& \xmark
& 0.8224
& 0.6593
& 15.3725
& 0.5318
& 0.3766 \\

& TSD-SR$^{*}$ (CVPR'25)
& \xmark
& 0.8238 
& 0.6599 
& 17.3098 
& 0.6165 
& 0.2849 \\
\cmidrule(lr){1-8}

\multirow{3}{*}{RefSR}
& ReFIR (NeurIPS'24)
& \cmark
& 0.8199 
& 0.6689 
& 15.5310 
& 0.5343 
& 0.4259 \\

& AdaRefSR (ICLR'26)
& \cmark
& 0.8311 
& 0.6858 
& 15.6200 
& 0.5629 
& 0.3523 \\

& AdaRefSR$^{*}$ (ICLR'26)
& \cmark
& 0.8435
& 0.6828
& 17.1132
& \underline{0.6260} 
& \underline{0.2842} \\
\cmidrule(lr){1-8}

\multirow{3}{*}{RefGCR}
& OmniPaint (ICCV'25)
& \cmark
& 0.7831 
& 0.5720 
& 13.4472 
& 0.5486 
& 0.5166 \\

& ImageCritic (CVPR'26)
& \cmark
& 0.8536 
& 0.7156 
& \underline{17.4391} 
& 0.6193
& 0.2991 \\

& ImageCritic$^{*}$ (CVPR'26)
& \cmark
& \underline{0.8542} 
& \underline{0.7165} 
& 17.2090 
& 0.6060 
& 0.3039 \\

\cmidrule(lr){1-8}

RefGC-SR$^2$
& Ours
& \cmark
& \textbf{0.8696} 
& \textbf{0.7474} 
& \textbf{17.5148} 
& \textbf{0.6335} 
& \textbf{0.2746} \\
\bottomrule
\end{tabular}
\end{table}
\begin{figure*}[t]
\centering
\begin{minipage}[t]{0.60\linewidth}
\vspace{0pt}
\centering
\tiny
\setlength{\tabcolsep}{2.5pt}
\sidecaptiontable{
In-the-wild evaluation, where outputs from compositing~\citep{huang2025dreamfuse,song2026insert} and customization~\citep{zhang2025freecus,feng2026personalize} models are treated as LRGI given HRRI and HRGT. \textbf{Compositing} denotes results using outputs from compositing models as LRGI, and \textbf{Customization} denotes results using outputs from customization models as LRGI.
}{tab:comp_custom_quant}
\begin{tabular}{lccccc@{\hspace{6pt}}cc}
\toprule
& \multicolumn{5}{c}{\textbf{Compositing}}
& \multicolumn{2}{c}{\textbf{Customization}} \\
\cmidrule(lr){2-6} \cmidrule(lr){7-8}
\textbf{Model}
& \textbf{CLIP-I$\uparrow$}
& \textbf{DINO$\uparrow$}
& \textbf{PSNR$\uparrow$}
& \textbf{SSIM$\uparrow$}
& \textbf{LPIPS$\downarrow$}
& \textbf{CLIP-I$^T\uparrow$}
& \textbf{DINO$^T\uparrow$} \\
\midrule
DiT4SR$^{*}$
& 0.8101
& 0.6057
& 13.6421
& 0.4781
& 0.4751
& 0.7184
& 0.3565 \\
TSD-SR$^{*}$
& 0.8188
& 0.5770
& 14.8691
& 0.4901
& \underline{0.4253}
& 0.7132
& 0.3160 \\
ReFIR
& 0.8026
& 0.5506
& 14.0969
& 0.4834
& 0.4805
& 0.7106
& 0.3260 \\
AdaRefSR$^{*}$
& 0.8290
& 0.6081
& 14.2231
& 0.4840
& 0.4341
& 0.7206
& 0.3406 \\
OmniPaint
& 0.7945
& 0.5570
& 14.1785
& 0.4873
& 0.4947
& 0.7106
& 0.3213 \\
ImageCritic$^{*}$
& \underline{0.8407}
& \underline{0.6446}
& \underline{14.9504}
& \underline{0.5045}
& 0.4335
& \underline{0.7240}
& \underline{0.3659} \\
Ours
& \textbf{0.8476}
& \textbf{0.6626}
& \textbf{15.1028}
& \textbf{0.5063}
& \textbf{0.4170}
& \textbf{0.7280}
& \textbf{0.3771} \\
\bottomrule
\end{tabular}
\end{minipage}%
\hfill%
\begin{minipage}[t]{0.36\linewidth}
\vspace{0pt}
\centering
\includegraphics[width=0.95\linewidth]{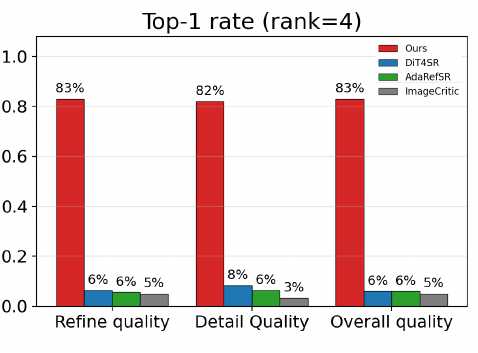}
\sidecaptionfigure{
Top-rank-1 user study results. The user study is conducted with 16 participants comparing one model from each SR, RefSR and RefGCR task and RefGC-SR$^2$ (Ours).
}{fig:user_study}
\end{minipage}
\vspace{-1em}
\end{figure*}

\paragraph{Implementation Details}  
\jhz{Details of the training setting on the RefGC-SR$^2$ Dataset and the evaluation protocols for Tables~\ref{tab:main_quan} and~\ref{tab:comp_custom_quant} are provided in Appendix~\ref{supsec:imp_detail}.}

\noindent\textbf{Training Datasets}
We denote the final triplet dataset generated using DipRefGC, described in Sec.~\ref{sec:DipRefGC}, as the RefGC-SR$^2$ Dataset. 
Our training set contains 40K triplets. 
\jhzz{For training, we downsample the images generated by DipRefGC to $256\times256$ using bicubic interpolation to obtain the LRGIs, while retaining the corresponding HRGT images at $1024\times1024$.}
% For LRGI, we downsample the original $1024\times1024$ (input HRGT resolution) images to $256\times256$ (output LRGI resolution) using bicubic interpolation.

\noindent\textbf{Evaluation Datasets}
We further propose the RefGC-SR$^2$ Benchmark for evaluation. 
This benchmark consists of 200 triplets generated from HRRI and HRGT pairs that are not included in the RefGC-SR$^2$ Dataset. 
In addition, to evaluate in-the-wild setting, we use extra HRRI and HRGT pairs that are not used in either the RefGC-SR$^2$ Dataset or Benchmark.
In-the-wild benchmark is constructed with HRRI-HRGT pairs that are not included in the RefGC-SR$^2$ Dataset or Benchmark and we generate the corresponding LRGIs by applying two recent compositing models, DreamFuse~\citep{huang2025dreamfuse} and InsertAnything~\citep{song2026insert}, and two customization models, FreeCus~\citep{zhang2025freecus} and PersonalizeAnything~\citep{feng2026personalize}. 
With 50 triplets, we obtain 200 samples after applying these four models and report the results for each RefGC task. \jhz{For customization, the background and target object often undergo substantial changes. Therefore, metrics requiring pixel-aligned comparison with the GT or region-based evaluation using the reference object location are not meaningful. We instead report full-image metrics, denoted as CLIP-I$^{T}$ and DINO$^{T}$.}
% As shown in Table~\ref{tab:comp_custom_quant}, our model achieves the best performance across all metrics, demonstrating the practical applicability of our RefGC-SR$^2$ model. \dah{See Appendix~\ref{subsec:independent_sources} for further evaluations.}

% \jhxx{Our \textbf{RefGC-SR$^2$ Dataset} consists of (LRGI, HRRI, HRGT) triplets, where HRRI and HRGT depict the same object instance under different views, and LRGI is a pose-aligned, artifact-containing low-resolution generation. HRRI-HRGT pairs are curated from three real-world high-resolution sources via VLM-based filtering~\citep{qwen3vl2025} and human verification. To synthesize pose-aligned LRGIs, we propose DipRefGC, a diptych-conditioned generator built on FLUX~\citep{flux1-dev} with dual ControlNets~\citep{zhang2023adding} that disentangle appearance (Inpainting)~\citep{controlnet-inpaint} and pose (Canny)~\citep{controlnet-canny} control. The final dataset contains 40K training triplets across 12 object categories. For evaluation, we construct \textbf{RefGC-SR$^2$ Benchmark}, consisting of 200 triplets that are disjoint from the training set. Further details on dataset construction are provided in Sec.~\ref{sec:dataset}.}
% \input{tables/main_quan}

\begin{figure*}[h!]
	\centering
	\includegraphics[width=0.8\linewidth]{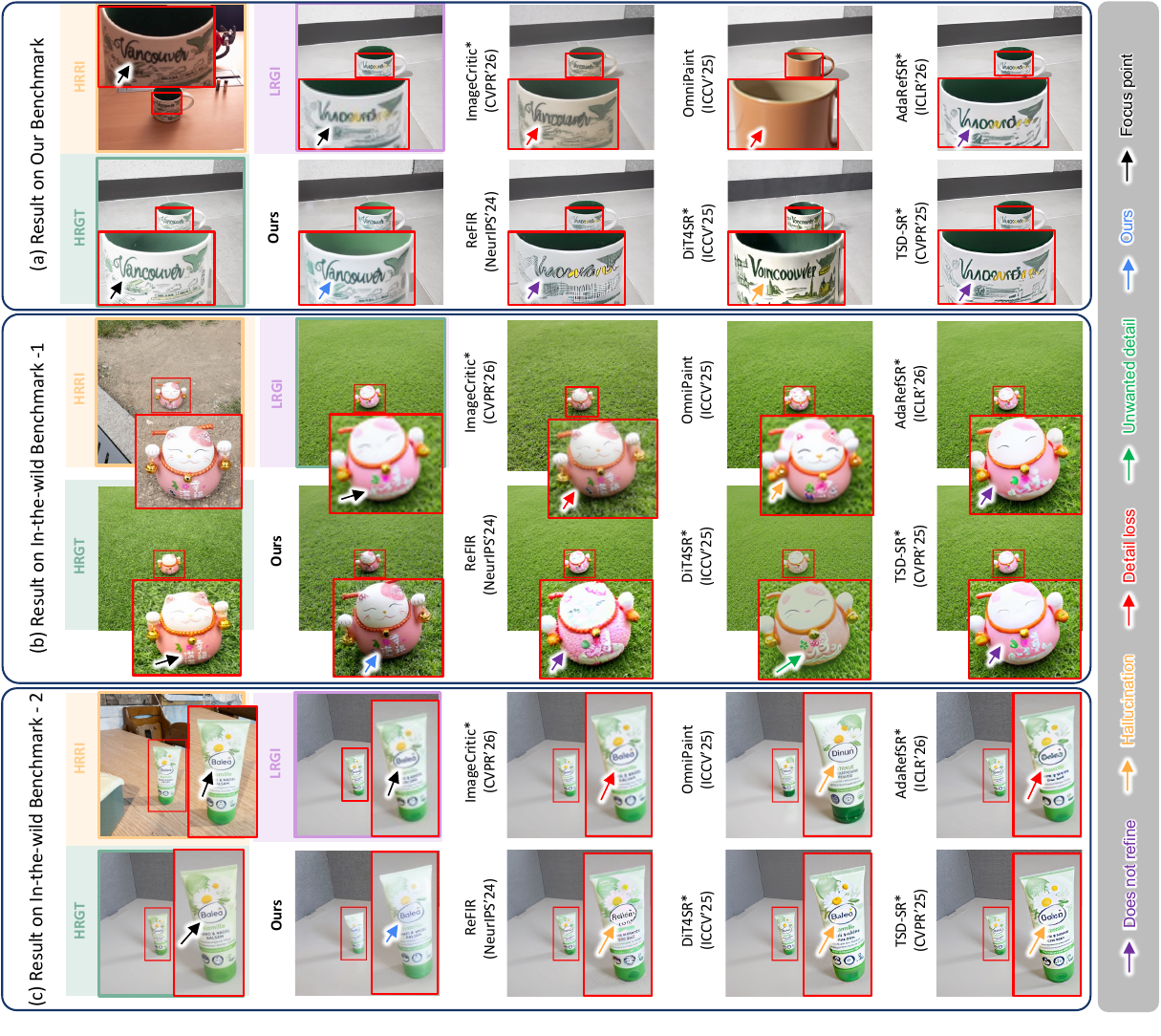}
    \caption{Qualitative comparison between our \jhmodel{} model and other models. 
    (a) shows qualitative results on the RefGC-SR$^2$ Benchmark, while (b) and (c) present results in the in-the-wild benchmark.
    Colored arrows mark failure cases of competing methods (see legend), and blue arrows indicate our results.
    Our \jhmodel{} model better preserves fine details from HRRI and achieves high-quality upscaling compared to competing methods. \textit{Best viewed in zoom.}}
    % \jh{Qualitative comparison between our \jhfix{\jhmodel{} model} and other models. 
    % (a) shows qualitative results on the RefGC-SR$^2$ Benchmark, while (b) and (c) present results in the in-the-wild setting.
    % \jhfix{Black arrows indicate key regions to focus on in the input. 
    % Blue arrows indicate results from our method. 
    % Green arrows mark unwanted details, red arrows denote detail loss, orange arrows highlight hallucinations, and purple arrows indicate regions where refinement fails.}
    % Our \jhfix{\jhmodel{} model} better preserves fine details from HRRI and achieves high-quality upscaling compared to competing methods.}
    \label{fig:main_qual}
% \vspace{-0.5em}
\end{figure*}

% \begin{figure*}[h!]
% 	\centering
% 	\includegraphics[width=0.8\linewidth]{figs/main_qual.pdf}
%     \caption{Qualitative comparison between our \jhmodel{} model and other models. 
%     (a) shows qualitative results on the RefGC-SR$^2$ Benchmark, while (b) and (c) present results in the in-the-wild benchmark.
%     Colored arrows mark failure cases of competing methods (see legend), and blue arrows indicate our results.
%     Our \jhmodel{} model better preserves fine details from HRRI and achieves high-quality upscaling compared to competing methods.}
%     % \jh{Qualitative comparison between our \jhfix{\jhmodel{} model} and other models. 
%     % (a) shows qualitative results on the RefGC-SR$^2$ Benchmark, while (b) and (c) present results in the in-the-wild setting.
%     % \jhfix{Black arrows indicate key regions to focus on in the input. 
%     % Blue arrows indicate results from our method. 
%     % Green arrows mark unwanted details, red arrows denote detail loss, orange arrows highlight hallucinations, and purple arrows indicate regions where refinement fails.}
%     % Our \jhfix{\jhmodel{} model} better preserves fine details from HRRI and achieves high-quality upscaling compared to competing methods.}
%     \label{fig:main_qual}
% \vspace{-0.5em}
% \end{figure*}

\noindent\textbf{Quantitative Results}
We conduct quantitative evaluation on our RefGC-SR$^2$ Benchmark and in-the-wild benchmark. 
Table~\ref{tab:main_quan} compares our \jhfix{\jhmodel{} model} with ImageCritic~\citep{ouyang2025consistency}, OmniPaint~\citep{yu2025omnipaint}, AdaRefSR~\citep{wang2026trust}, ReFIR~\citep{guo2024refir}, DiT4SR~\citep{duan2025dit4sr}, and TSD-SR~\citep{dong2025tsd}. 
Except for training-free methods~\citep{guo2024refir} or those without open-source training code~\citep{yu2025omnipaint}, we report additional results after fine-tuning on our training dataset for fair comparison (marked with * in Table~\ref{tab:main_quan}).
As shown in Table~\ref{tab:main_quan}, \jhzz{our} model achieves state-of-the-art performance across all metrics, demonstrating its effectiveness for the proposed RefGC-SR$^2$ task.
% We provide quantitative results on a in-the-wild benchmark in Table~\ref{tab:comp_custom_quant} to assess the generalization ability of our model to real-world applications. 
% \jhx{This experiment is conducted to simulate a practical setting, using outputs from real compositing and customization models.}
% In this experiment, we use 
To demonstrate the generalization ability of our model, we provide quantitative results on the in-the-wild benchmark described in Sec.~\ref{sec:experiment}-Dataset in Table~\ref{tab:comp_custom_quant}.
As shown in Table~\ref{tab:comp_custom_quant}, our model achieves the best performance across all metrics, demonstrating the practical applicability of our RefGC-SR$^2$ model.

% In-the-wild evaluation set is constructed with HRGT–HRRI pairs that are not included in the RefGC-SR$^2$ Dataset or Benchmark and we generate the corresponding LRGIs by applying two recent compositing models, DreamFuse~\citep{huang2025dreamfuse} and InsertAnything~\citep{song2026insert}, and two customization models, FreeCus~\citep{zhang2025freecus} and PersonalizeAnything~\citep{feng2026personalize}. 
% With 50 triplets, we obtain 200 samples after applying these four models and report the results for each RefGC task.
% As shown in Table~\ref{tab:comp_custom_quant}, our model achieves the best performance across all metrics, demonstrating the practical applicability of our RefGC-SR$^2$ model.
% \jhx{These results suggest that our RefGC-SR$^2$ model performs better than competing methods even in practical (in-the-wild) settings.}

\noindent\textbf{Qualitative Results} 
% % Fig.~\ref{fig:main_qual} presents qualitative results on RefGC-SR$^2$. 
% \jhxx{Fig.~\ref{fig:main_qual} presents qualitative results on both the proposed RefGC-SR$^2$ Benchmark and the practical in-the-wild benchmark. 
% We use colored arrows to highlight important regions, as described in the gray box on the right side of Fig.~\ref{fig:main_qual} (Please follow the arrows for detailed comparisons). 
% (a) shows results on the RefGC-SR$^2$ Benchmark, while (b) and (c) presents results in the in-the-wild setting. 
% As shown in these examples, our RefGC-SR$^2$ model effectively refines generative artifacts faithfully to HRRI (e.g., bring back identity loss) while successfully increasing image resolution. Please refer to the Appendix and supplementary material for additional qualitative results.}
\jhxx{Fig.~\ref{fig:main_qual} presents qualitative results on both the proposed RefGC-SR$^2$ Benchmark and the practical in-the-wild benchmark. 
We use colored arrows to highlight important regions, as described in the gray box on the right side of Fig.~\ref{fig:main_qual} (Please follow the arrows for detailed comparisons). 
(a) shows results on the RefGC-SR$^2$ Benchmark, while (b) and (c) present results in the in-the-wild setting. 
As shown in these examples, our RefGC-SR$^2$ model effectively refines generative artifacts faithfully to HRRI (e.g., bring back identity loss) while successfully increasing image resolution. 
We further provide additional qualitative results on other samples in the \textit{suppl.}, and present qualitative evaluations on the outputs of both commercial and open-source models in Appendix~\ref{sup:commercial_comp} to demonstrate the additional generalization capability of our RefGC-SR$^2$ model.}

\noindent\textbf{User study} 
To further demonstrate the advantage of our model, we conduct a user study under the in-the-wild setting. 
% We first generate LRGI by applying two compositing models~\citep{huang2025dreamfuse,song2026insert} and two customization models~\citep{zhang2025freecus,feng2026personalize}, all of which are reference-guided generative models. 
These outputs are then used as inputs for SR (DiT4SR~\citep{duan2025dit4sr}), RefSR (AdaRefSR~\citep{wang2026trust}), RefGCR (ImageCritic~\citep{ouyang2025consistency}), and RefGC-SR$^2$ (Ours) for comparison.
We evaluate the results using three criteria: refinement faithfulness (artifact removal and restoration based on HRRI), detail restoration (SR ability), and overall quality. 
\jhxx{As shown in Fig.~\ref{fig:user_study}, our model outperforms all others across all criteria, achieving top-1 rates of 83\%, 82\%, and 83\% on refine, detail, and overall quality, respectively, compared to at most 8\% for the competing methods, demonstrating its ability to effectively perform both refinement and super-resolution.} For more details on the user study protocol and results, please refer to \jhz{Appendix~\ref{supsec:user_study}}.
% As shown in Fig.~\ref{fig:user_study}, our model outperforms all others across all metrics, demonstrating its ability to effectively perform both refinement and super-resolution.}

\begin{figure*}[t]
\centering
\begin{minipage}[t]{0.52\linewidth}
\centering
\scriptsize
\setlength{\tabcolsep}{6pt}
\sidecaptiontable{
Ablation study of the proposed components on \jhmodel{} benchmark. 
The best and second-best results are highlighted in \textbf{bold} and \underline{underline}, respectively.
}{tab:ablation_main_model}
\begin{tabular}{lcccc}
\toprule
\textbf{Setting} & \textbf{(a)} & \textbf{(b)} & \textbf{(c)} & \textbf{(d)} \\
\midrule
\textbf{FreqMoLE} & \xmark & \xmark & \cmark & \cmark \\
\textbf{$\mathcal{L}_f$} & \xmark & \cmark & \xmark & \cmark \\
\midrule
\textbf{CLIP-I$\uparrow$} & 0.8437 & 0.8654 & \underline{0.8673} & \textbf{0.8696} \\
\textbf{DINO$\uparrow$} & 0.6870 & \underline{0.7386} & 0.7317 & \textbf{0.7474} \\
\textbf{PSNR$\uparrow$} & 16.3956 & 17.1981 & \underline{17.3893} & \textbf{17.5148} \\
\textbf{SSIM$\uparrow$} & 0.6068 & \underline{0.6221} & \underline{0.6221} & \textbf{0.6335} \\
\textbf{LPIPS$\downarrow$} & 0.3538 & \underline{0.2835} & 0.2896 & \textbf{0.2746} \\
\bottomrule
\end{tabular}
\end{minipage}%
\hfill%
\begin{minipage}[t]{0.46\linewidth}
\centering
\vspace{0pt}
\includegraphics[width=\linewidth]{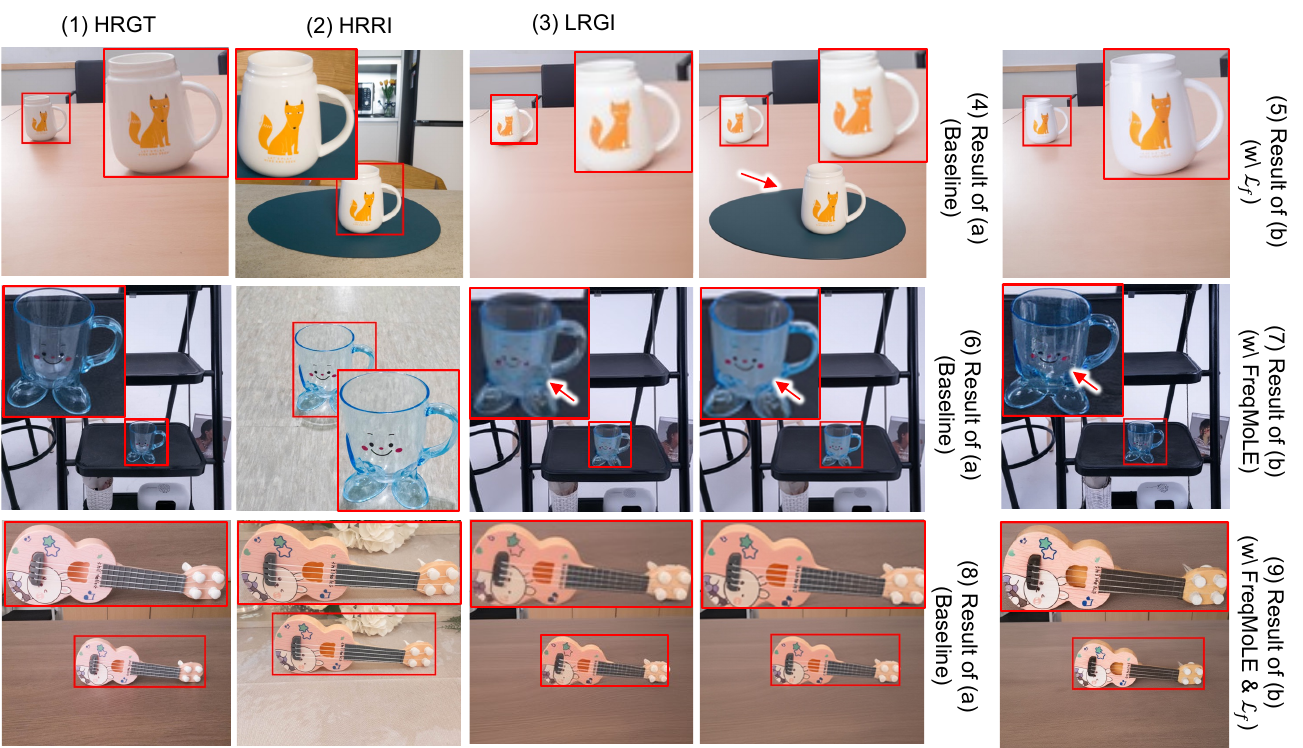}
\sidecaptionfigure{
Qualitative ablation results of the proposed model. \textit{Best viewed in~Zoom.}
}{fig:ablation_qual}
\end{minipage}
\end{figure*}

\noindent\textbf{Ablation Study on \jhfix{\jhmodel{} Model}}
Table~\ref{tab:ablation_main_model} reports the ablation results on RefGC-SR$^2$ Benchmark.
The baseline (a), removing both FreqMoLE and $\mathcal{L}_f$, shows the lowest performance across all metrics.
Adding $\mathcal{L}_f$ alone in (b) significantly improves identity preservation (DINO $+7.5\%$) and perceptual quality (LPIPS $-19.9\%$), confirming that decomposing supervision into LF alignment with HRGT and HF statistical matching with HRRI promotes faithful detail restoration.
This is also reflected in Fig.~\ref{fig:ablation_qual}: without $\mathcal{L}_f$ (4), HRRI is directly injected into the output, whereas with $\mathcal{L}_f$ (5), the model leverages HRRI while preserving the structure of HRGT.
Adding FreqMoLE alone in (c) improves reconstruction fidelity (PSNR $+0.99$ dB, CLIP-I $+2.8\%$), as layer-wise mixing of LF/HF experts aligns with the coarse-to-fine frequency hierarchy of FLUX-Kontext analyzed in Sec.~\ref{method:motivations}.
In Fig.~\ref{fig:ablation_qual}-(6), without FreqMoLE, the cup remains opaque (red arrow) and does not match the transparency in HRRI, while in (7) with FreqMoLE, the model correctly restores its transparency, consistent with both HRRI and HRGT.
% Combining both as in (d) achieves the best performance across all metrics, confirming that FreqMoLE and $\mathcal{L}_f$ play 
Combining both in (d) achieves the best performance across all metrics, confirming that FreqMoLE and $\mathcal{L}_f$ play complementary roles (structural modulation and band-specific supervision) that are both essential to \jhmodel{}.
As shown in Fig.~\ref{fig:ablation_qual}-(9), the final model effectively transfers fine details from HRRI while correcting mismatched regions such as the star and ear (red arrows). 
\jhz{Further analyses of high-frequency supervision and robustness to large viewpoint differences are provided in Appendix \ref{supsec:hf_supervision} and Appendix~\ref{supsec:large_viewpoint}, respectively.}

\noindent\textbf{Ablation Study on \dhmodel}
We further conduct an ablation study of DipRefGC to validate the role of pose conditioning and RefGC-oriented fine-tuning. 
The results show that Canny guidance improves pose consistency between LRGI and HRGT, while fine-tuning on both compositing and customization settings improves reference identity preservation and reduces the distribution gap to real RefGC outputs~\citep{feng2026personalize, zhang2025freecus, huang2025dreamfuse, song2026insert}. See Fig.~\ref{fig:diprefgc_triplet} for qualitative triplet examples. \dah{See Appendix~\ref{app:diprefgc_ablation} for further DipRefGC ablations.}
% \dh{Table~\ref{tab:diprefgc_ablation} analyzes the effect of each design choice in DipRefGC. The base DiptychPrompting model (a) shows limited structural alignment with the target HRGT, especially in customization. Adding Canny ControlNet (b) consistently improves mask IoU in both compositing and customization, indicating that edge-based guidance helps preserve pose and structural consistency between the synthesized LRGI and HRGT. 
% Joint fine-tuning further improves identity preservation across both RefGC modes, while also achieving the best FID in the compositing setting. Overall, the ablation shows that Canny ControlNet mainly contributes to pose consistency, whereas joint fine-tuning improves reference identity preservation and reduces the distribution gap to real RefGC outputs in the compositing setting.}

\section{Conclusion}
\label{sec:conclusion}
\jhx{In this paper, we propose \textbf{RefGC-SR$^2$}, a new task that jointly performs super-resolution and artifact refinement on generated images guided by HRRI. To support this task, we introduce the first real-world triplet dataset (RefGC-SR$^2$ Dataset) and the first dedicated model (RefGC-SR$^2$ Model). For dataset construction, we propose DipRefGC, a FLUX-based generator that synthesizes HRGT-aligned artifact-corrupted LRGIs, yielding triplets suitable for supervised training. After training on this dataset, our RefGC-SR$^2$ model is able to leverage HRRI to faithfully refine generative artifacts while recovering high-resolution details. Extensive experiments show that our model consistently outperforms SR, RefSR, and RefGCR baselines, preserving reference identity and restoring fine details, on our RefGC-SR$^2$ Benchmark as well as on real in-the-wild RefGC inputs.}

% \jhx{In this paper, we propose \textbf{RefGC-SR$^2$}, a new post-processing task that jointly performs super-resolution and artifact refinement for low-resolution generated images using the original HRRI. To support this task, we further introduce the first real-world triplet dataset (\textbf{RefGC-SR$^2$ Dataset}) and the first dedicated model (\textbf{RefGC-SR$^2$ Model}). Our work is motivated by the observation that, despite rapid progress in reference-guided super-resolution (RefSR) and reference-guided generated content refinement (RefGCR), neither addresses both objectives on generated content: RefSR assumes natural image degradation, while RefGCR is limited to fixed low resolution. To construct the dataset, we propose \textbf{DipRefGC}, a diffusion-based generator that disentangles appearance and pose control to synthesize artifact-corrupted LRGIs aligned with HRGT, yielding triplets suitable for supervised training. Built on this dataset, our \textbf{RefGC-SR$^2$ Model} integrates information from HRRI and HRGT to refine generative artifacts while recovering high-resolution details. Extensive experiments show that our model consistently outperforms SR, RefSR, and RefGCR baselines, preserving reference identity and restoring fine details.}

% \input{sec/6_ai_use}
% \input{sec/7_reproducibility_statement}

{
    \small
    \bibliographystyle{unsrtnat}
    \bibliography{main}
}

%%%%%%%%%%%%%%%%%%%%%%%%%%%%%%%%%%%%%%%%%%%%%%%%%%%%%%%%%%%%

\newpage
\appendix
\newpage
\section*{Appendix Contents}
\startcontents[appendix]
\printcontents[appendix]{}{1}{\setcounter{tocdepth}{2}}
\newpage

\section{Technical appendices and supplementary material}

We provide part of the qualitative results from the proposed RefGC-SR$^2$ Benchmark, along with additional comparisons, in the supplementary material. 
Due to space limitations, only a subset of the benchmark and comparison results is included. 
We plan to release the full qualitative results, RefGC-SR$^2$ Dataset, Benchmark, code, and checkpoints in the future.
% Technical appendices with additional results, figures, graphs, and proofs may be submitted with the paper submission before the full submission deadline (see above). You can upload a ZIP file for videos or code, but do not upload a separate PDF file for the appendix. There is no page limit for the technical appendices. 

% Note: Think of the appendix as ``optional reading'' for reviewers. The paper must be able to stand alone without the appendix; for example, adding critical experiments that support the main claims to an appendix is inappropriate. 
\section{Related Works}
\label{app:related_work}

\begin{figure}[t]
    \centering
    \includegraphics[width=0.9\linewidth]{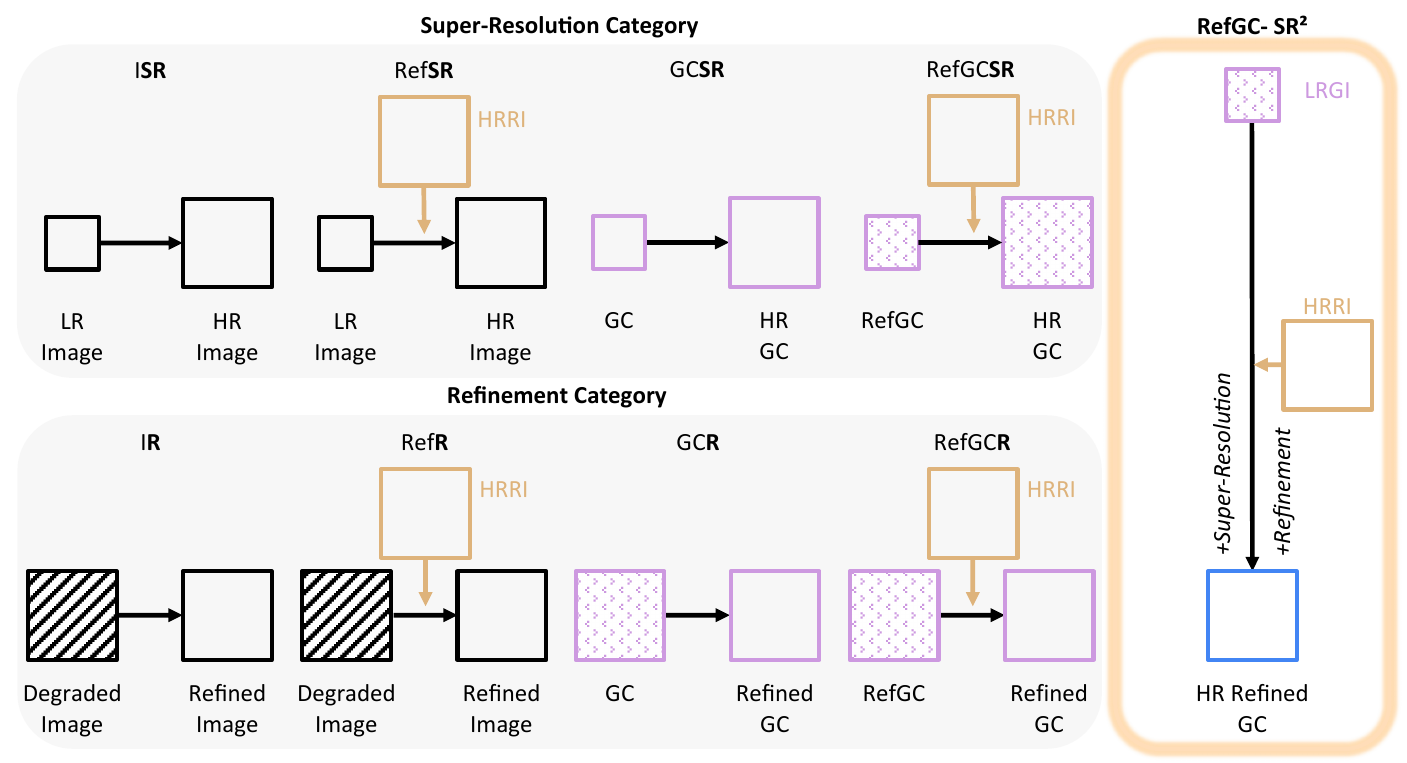}
    \caption{\textbf{Visual taxonomy of related image enhancement tasks.} We illustrate the input-output structure of all eight related tasks (Sec.~\ref{app:related_work}, Table~\ref{tab:task_comparison}) alongside our RefGC-SR$^2$. Tasks are organized into two categories: the \textit{Super-Resolution Category} (top), which enlarges spatial resolution, and the \textit{Refinement Category} (bottom), which removes artifacts or degradations at a fixed resolution. \textbf{Box size} denotes spatial resolution: smaller boxes are low-resolution (LR) inputs and larger boxes are high-resolution (HR) outputs, so a size change between input and output indicates super-resolution while equal sizes indicate resolution-preserving. \textbf{Box color} denotes the image type: \textit{black} for natural images, \textit{purple} for generated content (GC), \textit{yellow} for the user-provided high-resolution reference image (HRRI), and \textit{blue} for our final HR refined output. \textbf{Box pattern} denotes the degradation type: diagonal stripes indicate natural degradations such as blur and noise, while hatched fill patterns indicate generative artifacts such as identity distortion, detail inconsistency, texture loss, and overall quality degradation. Among all tasks, our RefGC-SR$^2$ is the only one that \emph{simultaneously} consumes LRGI and an HRRI as inputs and jointly performs super-resolution and generative-artifact refinement, producing an HR refined output.}
    \label{fig:tasks_taxonomy}
\end{figure}

\paragraph{Image Super-Resolution (ISR)}
Image Super-Resolution (ISR) aims to reconstruct a high-resolution (HR) image from a low-resolution (LR) observation. As shown in Fig.\ref{fig:tasks_taxonomy} (I\textbf{SR}), ISR takes a single LR image as input and produces an HR image, without using any external reference. Recent diffusion-based ISR methods\citep{choi2025framer, duan2025dit4sr, dong2025tsd} exploit pretrained generative priors and have achieved strong performance under natural-image degradation assumptions such as bicubic downsampling, blur, noise, compression, and realistic camera degradations. However, ISR is not designed for LR generated content images (LRGIs), whose degradations arise from generative pipelines rather than camera-side or hand-crafted degradation processes. Moreover, ISR cannot recover identity-specific information that is absent from the LRGI but still available in the user-provided HRRI. RefGC-SR$^2$ differs from ISR by explicitly targeting generated content with artifacts and by reusing the HRRI as an external recovery source.

\paragraph{Reference-guided Super-Resolution (RefSR)}
Reference-guided SR (RefSR) extends ISR by exploiting an external HR reference image (HRRI) to recover fine-grained details~\citep{lee2025reference, lee2024refqsr}. As shown in Fig.~\ref{fig:tasks_taxonomy} (Ref\textbf{SR}), RefSR takes an LR image and an HRRI as inputs, and produces an HR image. This formulation is related to ours in that it leverages a reference image as a source of high-frequency information. However, existing RefSR methods are still formulated for natural-image SR, where the LR input is assumed to be degraded from a real HR image. They do not model RefGC-specific artifacts such as identity distortion, detail inconsistency, hallucinated structures, or texture loss. Therefore, directly applying RefSR to LRGIs may sharpen the image but does not explicitly refine generative artifacts. RefGC-SR$^2$ inherits the reference-guided recovery concept of RefSR, but extends it to the generated-content domain and jointly performs generative artifact refinement.

\paragraph{Generated Content Super-Resolution (GCSR)}
Generated Content Super-Resolution (GCSR) aims to overcome the native resolution limits of generative models~\citep{jeong2025latent, du2024demofusion, tragakis2024one}. As shown in Fig.~\ref{fig:tasks_taxonomy} (GC\textbf{SR}), GCSR takes generated content as input and increases its resolution without using any HRRI. Existing GCSR methods rely on the internal priors of pretrained generative models to hallucinate missing details at higher resolution. This is effective when the goal is to upscale a generated image itself, but it is insufficient for reference-guided generated content (RefGC). In RefGC pipelines, the generated output may have already lost reference-specific identity, logo, texture, or material details due to low-resolution reference conditioning. Since GCSR has no access to the original HRRI, it cannot explicitly recover such missing reference information. In contrast, RefGC-SR$^2$ reintroduces the HRRI at the post-generation stage and uses it as an explicit source of high-frequency reference details.

\paragraph{Reference-guided Generated Content Super-Resolution (RefGCSR)}
We use RefGCSR to denote a possible setting that performs super-resolution on generated content with the help of an HRRI. As shown in Fig.~\ref{fig:tasks_taxonomy} (RefGC\textbf{SR}), RefGCSR would take both GC and HRRI as inputs and produce an HR generated output. However, this setting has not been systematically established as a dedicated task in the literature, and a straightforward implementation would be to apply existing RefSR methods to generated content. Such a formulation still focuses on resolution recovery and does not explicitly address generative artifacts inherited from RefGC pipelines. As a result, the output may become higher-resolution, but artifacts such as identity distortion, detail inconsistency, and texture loss can remain unresolved. RefGC-SR$^2$ is stricter and more comprehensive: it requires not only HRRI-guided super-resolution, but also generative-artifact refinement.

\paragraph{Image Restoration (IR)}
Image restoration (IR) maps a degraded image to a clean image by removing degradations such as noise, blur, compression artifacts, or other natural corruptions~\citep{ye2024learning}. As shown in Fig.~\ref{fig:tasks_taxonomy} (I\textbf{R}), IR refines a degraded natural image at the same spatial resolution, without using an HRRI. Although modern diffusion-based restoration methods provide strong natural-image priors, their assumptions differ from those of RefGC-SR$^2$. They do not consume generated content as a specific input domain, do not use the original HRRI, and do not perform super-resolution as part of the task. Therefore, IR cannot recover reference-specific details missing from LRGIs, nor can it produce an HR output. RefGC-SR$^2$ extends beyond IR by jointly using HRRI guidance, generated-content artifact modeling, and resolution recovery.

\paragraph{Generated Content Restoration (GCR)}
Generated Content Restoration (GCR) focuses on improving generated images by removing visible artifacts or low-quality regions~\citep{lin2024diffbir}. As shown in Fig.~\ref{fig:tasks_taxonomy} (GC\textbf{R}), GCR takes generated content as input and produces a refined generated image at the same resolution. This task is closer to ours than IR because it recognizes that generated images have a different artifact distribution from natural degraded images. However, GCR still operates without an HRRI. Consequently, it must infer missing details from the generated image alone, and cannot restore identity-specific information that was discarded during the upstream reference-guided generation process. Moreover, GCR is typically a fixed-resolution refinement problem. RefGC-SR$^2$ differs by using the HRRI as an external recovery source and by jointly performing super-resolution.

\paragraph{Reference-guided Refinement (RefR)}
Reference-guided refinement (RefR) uses an external reference image to refine a degraded image~\citep{zhang2024reference, guo2024refir}. As shown in Fig.~\ref{fig:tasks_taxonomy} (Ref\textbf{R}), RefR augments image restoration with an HRRI, but still focuses on natural-image degradations and fixed-resolution refinement. This setting shares with ours the use of an HRRI, but its target domain and output requirement are different. RefR is not formulated for generated content from RefGC pipelines, and it does not address the coupled problem of LR input, generative artifacts, and missing reference-specific details. RefGC-SR$^2$ extends reference-guided refinement to the RefGC setting and additionally requires HR output recovery.

\paragraph{Reference-guided Generated Content Refinement (RefGCR)}
Reference-guided Generated Content Refinement (RefGCR) is the most closely related task to ours. RefGCR methods~\citep{song2024refine, ouyang2025consistency, zhou2026refineanything, liu2025omnirefiner} use a reference image to correct visible artifacts or inconsistencies in generated content. As shown in Fig.~\ref{fig:tasks_taxonomy} (RefGC\textbf{R}), RefGCR consumes GC and HRRI as inputs, and produces a refined GC output. Representative methods perform local alignment, reference-guided detail correction, or region-specific refinement to improve consistency between generated content and the reference. However, existing RefGCR methods are primarily fixed-resolution refinement methods. They do not formulate the problem of recovering an HR output from an LR generated input. This distinction is critical for modern RefGC pipelines, where the user-provided HRRI is often downsampled before generation, causing fine-grained reference details to be discarded before the generated output is produced. RefGCR can refine artifacts at the current resolution, but cannot recover the lost HR information. RefGC-SR$^2$ resolves this limitation by treating artifact refinement and super-resolution as a single coupled post-generation task.

\section{Implementation Details}
\label{supsec:imp_detail}

\paragraph{RefGC-SR$^2$ Dataset}
\jhz{To construct the RefGC-SR$^2$ Dataset, we build DipRefGC on a frozen FLUX~\citep{flux1-dev} backbone with two LoRA-tuned~\citep{hu2022lora} ControlNets~\citep{controlnet-canny, controlnet-inpaint}. DipRefGC is trained on a single NVIDIA B200 GPU with a batch size of 16 and a learning rate of $10^{-4}$ for 19{,}000 iterations.}

\paragraph{RefGC-SR$^2$ Model}
\jhz{We build the RefGC-SR$^2$ Model on a frozen FLUX-Kontext~\citep{labs2025flux} backbone. For both DipRefGC and the RefGC-SR$^2$ Model, the VAE~\citep{kingma2013auto} and T5 encoders are kept frozen, and we follow ImageCritic~\citep{ouyang2025consistency} for the text encoder configuration. The RefGC-SR$^2$ Model is trained on a single NVIDIA B200 GPU for $4\times$ upscaling with a batch size of 16 and a learning rate of $10^{-4}$. The \jhzz{frequency-adaptive} loss ($\mathcal{L}_{f}$) weights are set to $\lambda_{LF}=0.1$ and $\lambda_{HF}=0.2$.}

\paragraph{In-the-wild Benchmark} \jhz{For compositing, CLIP-I and DINO are computed on the masked foreground object regions, while PSNR, SSIM, and LPIPS are computed on crops enclosing the object masks obtained using SAM3. For customization, we report only whole-image CLIP-I and DINO because customization models may synthesize the subject in a different layout, making region-restricted and pixel-aligned metrics unreliable. Therefore, the compositing and customization results follow separate evaluation protocols and are not directly comparable across the two settings.}

\section{Details on Auxiliary Loss Terms}
\label{sec:aux_losses}

The overall training objective of our RefGC-SR$^2$ Model combines our proposed \jhzz{frequency-adaptive} loss $\mathcal{L}_f$ (Sec.~\ref{sec:method}) with two auxiliary loss terms: the flow-matching loss $\mathcal{L}_{FM}$ inherited from the FLUX-Kontext backbone~\citep{labs2025flux}, and the attention alignment loss $\mathcal{L}_{aal}$ adopted from ImageCritic~\citep{ouyang2025consistency}. For completeness, we briefly review these two terms here.

\subsection{Flow-Matching Loss ($\mathcal{L}_{FM}$)}
\label{sec:flow_matching_loss}

FLUX-Kontext~\citep{labs2025flux} is trained under the rectified flow-matching framework, where the diffusion transformer is supervised to predict the velocity field that transports a Gaussian prior to the data distribution along a linear trajectory. Following this formulation, given a clean target latent $z_1$ (encoded from HRGT) and a Gaussian noise sample $z_0 \sim \mathcal{N}(0, I)$, we define a linear interpolant
\[
z_t = (1-t)\, z_0 + t\, z_1, \quad t \in [0, 1],
\]
whose ground-truth velocity along the path is $z_1 - z_0$. The model $v_\theta$ predicts the velocity at $z_t$ conditioned on the task inputs $c$ (the LRGI latent, the HRRI latent, and the text instruction). The flow-matching loss is then defined as
\[
\mathcal{L}_{FM} = \mathbb{E}_{t,\, z_0,\, z_1,\, c}\left[\,\big\| v_\theta(z_t, t, c) - (z_1 - z_0) \big\|_2^2\,\right].
\]
This is the same objective used when fine-tuning FLUX-Kontext with LoRA in standard adaptation pipelines, and it serves as the primary supervision signal that drives our RefGC-SR$^2$ Model to produce outputs aligned with HRGT.

\subsection{Attention Alignment Loss ($\mathcal{L}_{aal}$)}
\label{sec:attn_alignment_loss}

To encourage the model to faithfully transfer reference information from HRRI to the refined output, we adopt the attention alignment loss proposed by ImageCritic~\citep{ouyang2025consistency}. ImageCritic introduces a \emph{reference-guided attentive alignment} mechanism that supervises the cross-attention behavior between the model's output tokens and the HRRI tokens, encouraging the model to concentrate its attention on the object region of the HRRI rather than diffusing it across irrelevant background. This regularization is particularly important in our setting because the HRRI and HRGT differ in viewpoint and scene context, so the model must learn to attend to the object identity in the HRRI while ignoring its background. We adopt the same formulation as in the original ImageCritic work~\citep{ouyang2025consistency}, and refer the reader to it for the full mathematical specification.

\subsection{Overall Objective}
\label{sec:overall_obj}

Combining the two auxiliary terms above with our proposed $\mathcal{L}_f$, the full training objective is
\[
\mathcal{L} = \mathcal{L}_{FM} + \mathcal{L}_f + \mathcal{L}_{aal}.
\]
The three terms play complementary roles: $\mathcal{L}_{FM}$ provides the dominant generative supervision toward HRGT under the rectified-flow framework of FLUX-Kontext, $\mathcal{L}_{aal}$ controls \emph{where} the model attends in the HRRI, and our $\mathcal{L}_f$ (Sec.~\ref{sec:method}) controls \emph{which frequency bands} are transferred from each reference (low-frequency from HRGT, high-frequency from HRRI). This decomposition is consistent with the design of our FreqMoLE module, which routes information through frequency-specialized LoRA experts in a layer-depth-dependent manner.

\section{Dataset Statistics and Diversity}
\label{sec:dataset_statistic}

\dah{We provide additional statistics on the semantic coverage and geometric and photometric diversity of the RefGC-SR² Dataset. These statistics characterize the variation between the HRRI and the HRGT, which depict the same object instance under different capture conditions.}

\subsection{Category and Instance Coverage}

\dah{The RefGC-SR² Dataset contains 16 super-categories, 577 distinct object labels, and approximately 3,500 distinct object identities. Object labels are extracted from the source data with assistance
from Qwen3-VL~\citep{bai2025qwen3} and then normalized to obtain the final category counts. The 12 labels used during video collection describe broad source-level groups rather than the final semantic categories of the curated dataset. For example, the collection-stage label \textit{kitchen cookware} encompasses objects such as bottles, mugs, bowls, jars, and cups. These collection labels should therefore be distinguished from the category and instance counts of the final dataset.}

\begin{table}[hbt]
    \centering
    \caption{\dah{
    Geometric diversity and lighting-condition variation of HRRI--HRGT pairs. Values are reported as median [interquartile range]. Viewpoint statistics are computed from a stratified sample of 3,000 pairs, while scale and foreground-color statistics use all 40,000 valid pairs. Foreground-color distance is computed as the Euclidean distance between mean foreground colors in 8-bit encoded LAB space.
    }}
    \label{tab:dataset_geometric_photometric_diversity}
    \small
    \setlength{\tabcolsep}{5pt}
    \begin{tabular}{lccc}
        \toprule
        Source
        & \shortstack{Viewpoint difference\\$\Delta$ (degrees)}
        & \shortstack{Scale ratio\\$\sqrt{a_{\mathrm{HRGT}}/a_{\mathrm{HRRI}}}$}
        & \shortstack{Foreground-color distance\\(encoded LAB)} \\
        \midrule
        uCO3D
        & \textbf{82.9} [54.8--133.1]
        & 1.00 [0.90--1.11]
        & 32.0 [16--62] \\
        ORIDa
        & 29.8 [12.1--149.9]
        & 1.00 \textbf{[0.68--1.47]}
        & \textbf{88.0} [46--166] \\
        \bottomrule
    \end{tabular}
\end{table}

\subsection{Geometric Diversity}

\dah{We quantify HRRI--HRGT variation in the 40K training pairs from ORIDa~\citep{kim2025orida} and uCO3D~\citep{liu2025uncommon} using three measures. Viewpoint difference is the relative camera rotation estimated with MASt3R~\citep{leroy2024grounding} on foreground object regions. The scale ratio is defined as $\sqrt{a_{\mathrm{HRGT}}/a_{\mathrm{HRRI}}}$, where $a=A/(WH)$ is the foreground mask area normalized by the corresponding image area. As a proxy for differences in lighting conditions and other photometric changes, we compute the Euclidean distance between the mean foreground color vectors of HRRI and HRGT in 8-bit encoded LAB space.}

\dah{For viewpoint estimation, we draw a stratified sample of 3,000 pairs: 750 ORIDa compositing pairs, 750 ORIDa customization pairs, and 1,500 uCO3D pairs. Scale and foreground-color statistics use 40,000 valid pairs, comprising 20,000 ORIDa pairs and 20,000 uCO3D pairs.}

\dah{Table~\ref{tab:dataset_geometric_photometric_diversity} reports the median and interquartile range for each measure. uCO3D exhibits a larger median viewpoint difference, whereas ORIDa exhibits wider scale variation and larger foreground-color differences. Although both sources have a median scale ratio of 1.00, their interquartile ranges show variation in individual pairs. These statistics demonstrate the diversity of HRRI--HRGT pairs in viewpoint, scale, and lighting conditions.}

\section{DipRefGC Ablations and LRGI Validation}
\label{app:diprefgc_ablation}

%==========================================================
\subsection{Design and Training Ablations}
\label{subsec:diprefgc_ablation}
\begin{table*}[hbt]
\centering
\caption{
    \dah{Ablation of DipRefGC on the RefGC-SR$^2$ Benchmark. We evaluate identity preservation using DINOv2~\citep{oquab2023dinov2} and CLIP-I~\citep{radford2021learning} similarity to the HRRI, pose consistency using object-mask IoU with the HRGT, and image-feature distribution similarity using FID~\citep{heusel2017gans} against the outputs of external RefGC generators. Rows (a)--(e) report the core ablation, and the shaded row (e) is our final model. Row (f) further fine-tunes (e) with extended data to examine the potential benefit of additional training data. Bold and underline indicate the best and second-best distinct values in each column, respectively; ties share the same formatting.}
}
\label{tab:diprefgc_ablation}
\tiny
\setlength{\tabcolsep}{4.2pt}
\begin{tabular}{l | cccc | cccc}
\toprule
& \multicolumn{4}{c|}{\textbf{Compositing}}
& \multicolumn{4}{c}{\textbf{Customization}}\\
\cmidrule(lr){2-5}\cmidrule(lr){6-9}
Variant
& DINO $\uparrow$ & CLIP-I $\uparrow$ & IoU $\uparrow$ & FID $\downarrow$
& DINO $\uparrow$ & CLIP-I $\uparrow$ & IoU $\uparrow$ & FID $\downarrow$\\
\midrule
(a) \quad DiptychPrompting~\citep{shin2025large} (base)
& 0.539 & 0.823 & 0.480 & 106.92
& 0.484 & 0.796 & 0.311 & \textbf{104.94}\\
(b) \quad DiptychPrompting + Canny
& 0.622 & 0.841 & 0.601 & 99.34
& 0.634 & 0.845 & 0.601 & \underline{107.64}\\
\midrule
\multicolumn{9}{l}{\emph{Phase 1}}\\
(c) \quad Compo-only
& \textbf{0.691} & \textbf{0.863} & \textbf{0.641} & 97.52
& 0.680 & \underline{0.854} & \textbf{0.630} & 109.18\\
(d) \quad Custom-only
& 0.664 & 0.854 & 0.624 & 100.45
& \underline{0.683} & \textbf{0.856} & \underline{0.621} & 109.69\\
\rowcolor{gray!10}
(e) \quad Jointly fine-tuned (final)
& 0.684 & 0.860 & \underline{0.637} & \underline{97.24}
& 0.677 & 0.853 & 0.602 & 109.27\\
\midrule
\multicolumn{9}{l}{\emph{Phase 1 $\to$ Phase 2}}\\
(f) \quad + Extended data fine-tuning
& \underline{0.687} & \underline{0.861} & \textbf{0.641} & \textbf{95.97}
& \textbf{0.684} & \textbf{0.856} & 0.605 & 107.91\\
\bottomrule
\end{tabular}
\end{table*}

\dah{Table~\ref{tab:diprefgc_ablation} reports the ablation study of DipRefGC. Adding Canny ControlNet~\citep{controlnet-canny} (b) improves mask IoU over the base DiptychPrompting model~\citep{shin2025large} from 0.480 to \textbf{0.601} in compositing and from 0.311 to \textbf{0.601} in customization, verifying its role in enforcing the pose constraint between the LRGI and HRGT. RefGC-oriented fine-tuning improves identity preservation, with higher DINO and CLIP-I
scores than the Canny-only variant in both compositing-only and customization-only settings. Joint fine-tuning provides a practical balance across the two RefGC modes, achieving competitive identity and pose scores while improving FID in compositing.}

For training, Phase~1 uses 40K triplets constructed from ORIDa~\citep{kim2025orida} and uCO3D~\citep{liu2025uncommon}, consisting of 20K samples from each dataset. \dah{Phase~2 further fine-tunes the final model (e) on 12K additional triplets from UltraVideo~\citep{xue2025ultravideo}, which contains more challenging videos with richer subject motion.} This additional fine-tuning reduces FID in both compositing and customization, suggesting that DipRefGC can benefit from more challenging motion-rich data while preserving the pose consistency required for RefGC-SR$^2$ training.

%==========================================================
\subsection{Validation of DipRefGC-Generated LRGIs}
\label{subsec:RefGC_artifact}

\dah{We further examine whether the LRGIs generated by DipRefGC exhibit artifact characteristics similar to those of outputs from existing RefGC generators.}

\paragraph{\dah{Evaluation Setup}}
\dah{All comparisons use the 200 HRRI--HRGT pairs of the RefGC-SR$^2$ Benchmark. For every benchmark pair, DipRefGC generates one compositing LRGI and one customization LRGI. As external generators, we use DreamFuse~\citep{huang2025dreamfuse} and InsertAnything~\citep{song2026insert} for compositing, and FreeCus~\citep{zhang2025freecus} and PersonalizeAnything~\citep{feng2026personalize} for customization; each external generator also produces one output for every benchmark pair.}

\dah{We characterize four aspects of RefGC artifacts. Identity similarity is measured by DINOv2~\citep{oquab2023dinov2} cosine similarity to the HRRI on foreground object crops. Detail similarity is measured by masked SSIM~\citep{wang2004image} against the HRGT, averaging the SSIM map over the HRGT foreground mask. Texture preservation is measured by the high-frequency energy ratio relative to the HRGT: within the crop defined by the HRGT object mask, we compute the fraction of Fourier power above a normalized radial frequency of 0.25 and divide the output fraction by the corresponding HRGT fraction. Overall quality is assessed on the full output image by Qwen3-VL-8B-Instruct~\citep{bai2025qwen3}, which assigns an integer score from 0 to 4, with higher scores indicating more pronounced artifacts.}

\dah{For each metric, we compute the one-dimensional 1-Wasserstein distance~\citep{ramdas2017wasserstein} between two empirical score distributions, where lower values indicate closer distributions. DINOv2 cosine similarity and SSIM lie in $[-1,1]$, so the corresponding distances lie in $[0,2]$. The VLM score lies in $[0,4]$, giving distances in $[0,4]$. The high-frequency energy ratio is nonnegative and has no fixed upper bound. Because the metrics have different scales, distances are interpreted only within each metric.}

%----------------------------------------------------------
\begin{table}[t]
    \centering
    \caption{\dah{Comparison of artifact metric distributions on the 200-pair RefGC-SR$^2$ Benchmark. We compare 200 outputs from the final DipRefGC model with 200 external outputs, selecting 50 per external generator and one DipRefGC output for the same pair and task. Each benchmark pair appears once in each pool. We report the one-dimensional 1-Wasserstein distance between the pooled distributions and the mean and maximum of the six pairwise distances among the four external output sets. Lower values indicate closer distributions within each metric.}}
    \label{tab:artifact_distribution}
    \footnotesize
    \setlength{\tabcolsep}{6pt}
    \begin{tabular}{lcc}
        \toprule
        Metric (score range)
        & \shortstack{DipRefGC vs.\\pooled external outputs}
        & \shortstack{Among external generators\\Mean / Max} \\
        \midrule
        Identity similarity (DINO, $[-1,1]$)
        & \textbf{0.0542} & 0.2139 / 0.3649 \\
        Detail similarity (mSSIM, $[-1,1]$)
        & 0.0716 & 0.0649 / 0.1173 \\
        Texture preservation (HF ratio, $\geq 0$)
        & \textbf{0.2406} & 0.3916 / 0.5186 \\
        Overall quality (VLM, $[0,4]$)
        & \textbf{0.1600} & 0.3400 / 0.5200 \\
        \bottomrule
    \end{tabular}
\end{table}

\paragraph{\dah{Comparison with Existing RefGC Generators}}
\dah{Table~\ref{tab:artifact_distribution} compares the final DipRefGC model (e) with the external generators on the benchmark. We partition the 200 benchmark pairs into four disjoint groups of 50, stratified by data source with a fixed random seed, and assign one group to each external generator. This yields 50 outputs per external generator and 200 external outputs in total. For each assigned pair, we take the DipRefGC output of the same task, giving 200 DipRefGC outputs (100 compositing and 100 customization). Both pools therefore cover every benchmark pair exactly once with the same task composition. As a descriptive reference for variation among existing generators, we report the mean and maximum of the six pairwise distances among the four external output sets. Because these sets are defined on disjoint pairs, the reference also reflects differences in content and sampling, and we use it for descriptive comparison rather than as a statistical threshold.}

\dah{The distances between DipRefGC and the pooled external outputs are \textbf{0.0542}, 0.0716, \textbf{0.2406}, and \textbf{0.1600} for identity similarity, detail similarity, texture preservation, and overall artifact severity, respectively. The identity, texture, and overall-artifact distances are smaller than the corresponding mean distances among the external generators (0.2139, 0.3916, and 0.3400). The detail distance of 0.0716 exceeds the corresponding mean of 0.0649 but remains below the maximum of 0.1173. Thus, on three of the four measured dimensions, DipRefGC outputs differ from the external outputs by less than the external generators typically differ from one another, and on the remaining dimension the difference stays within the observed range among them.}

\section{Further Analysis and Results of Our RefGC-SR$^2$ Model}
\label{supsec:further_analysis}
\jhz{In this section, we provide additional analyses and experimental results for our RefGC-SR$^2$ model to further examine its design choices and robustness under various conditions.}

\subsection{Analysis of High-Frequency Supervision}
\label{supsec:hf_supervision}

\jhz{In this section, we present the motivation for high-frequency supervision and provide empirical evidence supporting its design. The HRRI and HRGT depict the same object but are generally not spatially registered because of differences in viewpoint, pose, and object placement. Direct pixel-wise supervision against the HRRI can therefore impose incorrect constraints between spatial locations that do not correspond to each other. To avoid this problem, our method retains the HRGT-anchored flow-matching and low-frequency objectives, $\mathcal{L}_{\mathrm{FM}}$ and $\mathcal{L}_{\mathrm{LF}}$, while using $\mathcal{L}_{\mathrm{HF}}$ as an auxiliary distributional constraint.}

\jhz{
% We conduct an ablation using our RefGC-SR$^2$ model to examine four supervision strategies. 
\jhzz{We conduct an ablation using the RefGC-SR$^2$ model to examine four supervision strategies, as summarized in Table~\ref{tab:hf_supervision_ablation}.}
Setting (a) removes $\mathcal{L}_{\mathrm{f}}$ and serves as the baseline for measuring its contribution. Setting (b) replaces $\mathcal{L}_{\mathrm{HF}}$ with a pixel-wise $\ell_1$ loss against the HRRI to test whether direct reference supervision is effective. Setting (c) applies a full-band $\ell_1$ loss against the HRGT to determine whether HRGT-only reconstruction supervision is sufficient. Setting (d) is our proposed approach, which matches the channel-wise mean and standard deviation of the high-frequency components between the output and HRRI. All settings use the same FreqMoLE backbone, training schedule, Gaussian frequency decomposition and $\mathcal{L}_{\mathrm{FM}}$.}

\begin{table}[hbt]
\centering
\caption{Ablation of \jhzz{frequency-adaptive} supervision in the RefGC-SR$^2$ model. All settings share the same FreqMoLE backbone, training schedule, and flow-matching loss $\mathcal{L}_{\mathrm{FM}}$, while varying the auxiliary supervision. The best and second-best results are highlighted in \textbf{bold} and \underline{underline}, respectively.}
\label{tab:hf_supervision_ablation}
\setlength{\tabcolsep}{4pt}
\renewcommand{\arraystretch}{1.05}
\resizebox{0.70\textwidth}{!}{%
\begin{tabular}{lccccc}
\toprule
Auxiliary supervision
& CLIP-I
& DINO
& PSNR
& SSIM
& LPIPS \\
\midrule
(a) None (w/o $\mathcal{L}_f$)
& \underline{0.8673}
& 0.7317
& 17.3893
& \underline{0.6221}
& \underline{0.2896} \\

(b) Pixel-wise $\ell_1$ to HRRI
& 0.8656
& \underline{0.7453}
& 16.9123
& 0.6202
& 0.2960 \\

(c) Full-band $\ell_1$ to HRGT
& 0.8578
& 0.7243
& \textbf{17.7273}
& 0.6191
& 0.2929 \\

(d) Distributional $\mathcal{L}_{\mathrm{HF}}$ to HRRI (Ours)
& \textbf{0.8696}
& \textbf{0.7474}
& \underline{17.5148}
& \textbf{0.6335}
& \textbf{0.2746} \\
\bottomrule
\end{tabular}
}
\end{table}

\jhz{
Setting (a) provides a strong reconstruction baseline without $\mathcal{L}_f$. Setting (b) improves DINO over (a), but degrades CLIP-I and all reconstruction metrics, suggesting that pointwise reference supervision can impair reconstruction fidelity when the HRRI and HRGT are not spatially aligned. Setting (c) achieves the highest PSNR, but produces the lowest CLIP-I and DINO scores. Its SSIM and LPIPS are also worse than those of (a). This indicates that full-band HRGT supervision encourages pixel fidelity but does not explicitly associate the output high-frequency content with reference-specific details from the HRRI.
Setting (d) achieves the best CLIP-I, DINO, SSIM, and LPIPS scores while maintaining the second-best PSNR. Its improvements over (a) support the effectiveness of the proposed \jhzz{frequency-adaptive} supervision. Although channel-wise statistics admit multiple spatial configurations, this many-to-one property applies only to the auxiliary regularizer. In setting (d), the output remains anchored to the HRGT through $\mathcal{L}_{\mathrm{FM}}$ and $\mathcal{L}_{\mathrm{LF}}$, allowing reference-based high-frequency supervision without imposing invalid pixel-level correspondence.
}

% \jhz{Setting (d) achieves the best results on CLIP-I, DINO, SSIM, and LPIPS while maintaining the second-best PSNR. Although channel-wise statistics admit multiple spatial configurations, this many-to-one property applies only to the auxiliary regularizer. The output remains anchored to the HRGT through $\mathcal{L}_{\mathrm{FM}}$ and $\mathcal{L}_{\mathrm{LF}}$. These results demonstrate that the proposed distributional supervision incorporates reference-specific high-frequency details without imposing invalid pixel-level correspondence.}

\subsection{Robustness to Large Viewpoint Samples}
\label{supsec:large_viewpoint}
\jhz{In this section, we further evaluate whether RefGC-SR$^2$ can effectively leverage information from the HRRI under substantial HRRI-HRGT viewpoint discrepancies.}
\jhz{Since the HRRI is an independent observation of the same object instance, substantial viewpoint differences can arise between the HRRI and HRGT. We quantify this variation by estimating the relative HRRI-to-HRGT camera rotation $\Delta$ using MASt3R~\citep{leroy2024grounding} on the foreground object regions. The median rotation across the RefGC-SR$^2$ Benchmark is $72^\circ$. Moreover, $55\%$, $41\%$, and $22.5\%$ of the image pairs exhibit rotations greater than $60^\circ$, $90^\circ$, and $135^\circ$, respectively. These statistics indicate that the benchmark already contains substantial viewpoint variation.
For a more challenging evaluation, we select the $25\%$ of benchmark samples with the largest viewpoint differences. This subset contains 50 samples satisfying $\Delta \geq 132.5^\circ$. Table~\ref{tab:large_viewpoint} presents the quantitative results on this subset.}

\begin{table}[t]
\centering
\caption{Quantitative comparison on the large-viewpoint subset of the RefGC-SR$^2$ Benchmark. The subset contains the $25\%$ of samples with the largest HRRI-HRGT viewpoint differences, corresponding to $\Delta \geq 132.5^\circ$ and $N=50$. The best and second-best results are highlighted in \textbf{bold} and \underline{underline}, respectively. An asterisk ($*$) indicates a baseline finetuned on the RefGC-SR$^2$ Dataset.}
\label{tab:large_viewpoint}
\begingroup
\small
\setlength{\tabcolsep}{5pt}
\renewcommand{\arraystretch}{1.0}
\resizebox{0.60\textwidth}{!}{%
\begin{tabular}{lccccc}
\toprule
Model
& CLIP-I$\uparrow$
& DINO$\uparrow$
& PSNR$\uparrow$
& SSIM$\uparrow$
& LPIPS$\downarrow$ \\
\midrule
OmniPaint
& 0.7999
& 0.5870
& 12.99
& 0.5365
& 0.4958 \\

DiT4SR$^*$
& 0.8291
& 0.6529
& 15.03
& 0.5034
& 0.4349 \\

TSD-SR$^*$
& 0.8199
& 0.6277
& 15.70
& 0.4490
& 0.4627 \\

ReFIR
& 0.8252
& 0.6518
& 15.58
& 0.5457
& 0.4149 \\

AdaRefSR$^*$
& 0.8355
& 0.6504
& \textbf{16.56}
& \underline{0.5918}
& \textbf{0.3144} \\

ImageCritic$^*$
& \underline{0.8444}
& \underline{0.6755}
& 15.70
& 0.5854
& 0.3964 \\

RefGC-SR$^2$ (Ours)
& \textbf{0.8742}
& \textbf{0.7451}
& \underline{16.32}
& \textbf{0.6009}
& \underline{0.3196} \\
\bottomrule
\end{tabular}%
}
\endgroup
\end{table}

\jhz{RefGC-SR$^2$ achieves the best CLIP-I, DINO, and SSIM scores, together with the second-best PSNR and LPIPS. Compared with its performance on the complete benchmark in Table~\ref{tab:main_quan}, CLIP-I increases by $0.0046$, while DINO decreases by only $0.0023$. In addition, its DINO margin over the strongest baseline increases from $0.0309$ on the complete benchmark to $0.0696$ on the large-viewpoint subset. These results demonstrate that RefGC-SR$^2$ maintains strong identity preservation and reconstruction quality under substantial viewpoint discrepancies. If the reference depicts a completely different side of the object, appearance information that is entirely absent from the HRRI cannot be faithfully recovered. This limitation is inherent to reference-guided restoration methods.}

\subsection{Sensitivity to Reference Quality}
\label{supsec:reference_quality}

\jhzz{We evaluate how degradation of the HRRI affects refinement and super-resolution performance. Although the RefGC-SR$^2$ task assumes a high-quality, high-resolution reference image, references available in practical applications may be degraded. We consider one condition for each of three degradation types: Gaussian blur with $\sigma=1$, JPEG compression with $q=15$, and $4\times$ spatial downsampling from $1024\times1024$ to $256\times256$.}

\paragraph{Quantitative Results}
\jhzz{We evaluate a randomly selected subset of 100 samples. Each degradation is applied only to the HRRI, while the LRGI remains unchanged. All methods use the same samples, masks, and reference conditions, and none is retrained for the degraded-reference conditions. We report DINO for identity preservation, PSNR for reconstruction fidelity, and LPIPS for perceptual similarity. Values in parentheses in Table~\ref{tab:reference_quality} indicate the change relative to the clean-reference result of the same model.}
\jhzz{The results in Table~\ref{tab:reference_quality} show that HRRI degradation affects performance even when the LRGI is unchanged. For RefGC-SR$^2$, DINO decreases by 0.001 to 0.040, PSNR decreases by 0.83 to 1.63 dB, and LPIPS increases by 0.026 to 0.054 across the three conditions. These changes indicate that degraded reference information can adversely affect identity preservation, reconstruction fidelity, and perceptual similarity. The magnitude and direction of the changes vary across models and metrics, including a slight LPIPS improvement for AdaRefSR under blur. Our model achieves the highest DINO score under all three degraded conditions, but its performance remains affected by HRRI degradation.}

\begin{table}[t]
\centering
\caption{Effects of HRRI degradation on a randomly selected subset of 100 samples from the RefGC-SR$^2$ Benchmark. Only the HRRI is degraded, and no model is retrained. Parentheses show the change relative to the same model under the clean condition ($\Delta=\mathrm{degraded}-\mathrm{clean}$). Negative changes in DINO and PSNR and positive changes in LPIPS indicate worse performance. The best and second-best metric values within each condition are shown in \textbf{bold} and with \underline{underlining}, respectively. An asterisk ($*$) denotes a baseline fine-tuned on the RefGC-SR$^2$ Dataset.}
\label{tab:reference_quality}
\begingroup
\small
\setlength{\tabcolsep}{5pt}
\renewcommand{\arraystretch}{1.08}
\begin{tabular}{llccc}
\toprule
HRRI condition
& Model
& DINO$\uparrow$
& PSNR$\uparrow$
& LPIPS$\downarrow$ \\
\midrule

\multirow{3}{*}{Clean}
& AdaRefSR$^*$
& 0.692
& 17.34
& 0.289 \\
& ImageCritic$^*$
& \underline{0.742}
& \underline{17.62}
& \underline{0.285} \\
& RefGC-SR$^2$ (Ours)
& \textbf{0.751}
& \textbf{18.54}
& \textbf{0.256} \\
\midrule

\multirow{3}{*}{Blur}
& AdaRefSR$^*$
& 0.688 $(-0.004)$
& \underline{17.09} $(-0.25)$
& \underline{0.287} $(-0.002)$ \\
& ImageCritic$^*$
& \underline{0.697} $(-0.045)$
& 16.11 $(-1.51)$
& 0.444 $(+0.159)$ \\
& RefGC-SR$^2$ (Ours)
& \textbf{0.750} $(-0.001)$
& \textbf{17.69} $(-0.85)$
& \textbf{0.282} $(+0.026)$ \\
\midrule

\multirow{3}{*}{JPEG}
& AdaRefSR$^*$
& 0.665 $(-0.027)$
& \underline{17.15} $(-0.19)$
& \underline{0.299} $(+0.010)$ \\
& ImageCritic$^*$
& \underline{0.688} $(-0.054)$
& 16.13 $(-1.49)$
& 0.428 $(+0.143)$ \\
& RefGC-SR$^2$ (Ours)
& \textbf{0.729} $(-0.022)$
& \textbf{17.71} $(-0.83)$
& \textbf{0.283} $(+0.027)$ \\
\midrule

\multirow{3}{*}{\shortstack[l]{downsampling}}
& AdaRefSR$^*$
& 0.684 $(-0.008)$
& \textbf{16.98} $(-0.36)$
& \textbf{0.290} $(+0.001)$ \\
& ImageCritic$^*$
& \underline{0.695} $(-0.047)$
& 16.00 $(-1.62)$
& 0.461 $(+0.176)$ \\
& RefGC-SR$^2$ (Ours)
& \textbf{0.711} $(-0.040)$
& \underline{16.91} $(-1.63)$
& \underline{0.310} $(+0.054)$ \\
\bottomrule
\end{tabular}
\endgroup
\end{table}

% Before using this figure and caption, replace the existing
% 8x-downsampling results in rows (c) and (d) with actual 4x results.
\begin{figure*}[p]
  \centering
  \includegraphics[
    width=\linewidth,
    height=0.86\textheight,
    keepaspectratio
  ]{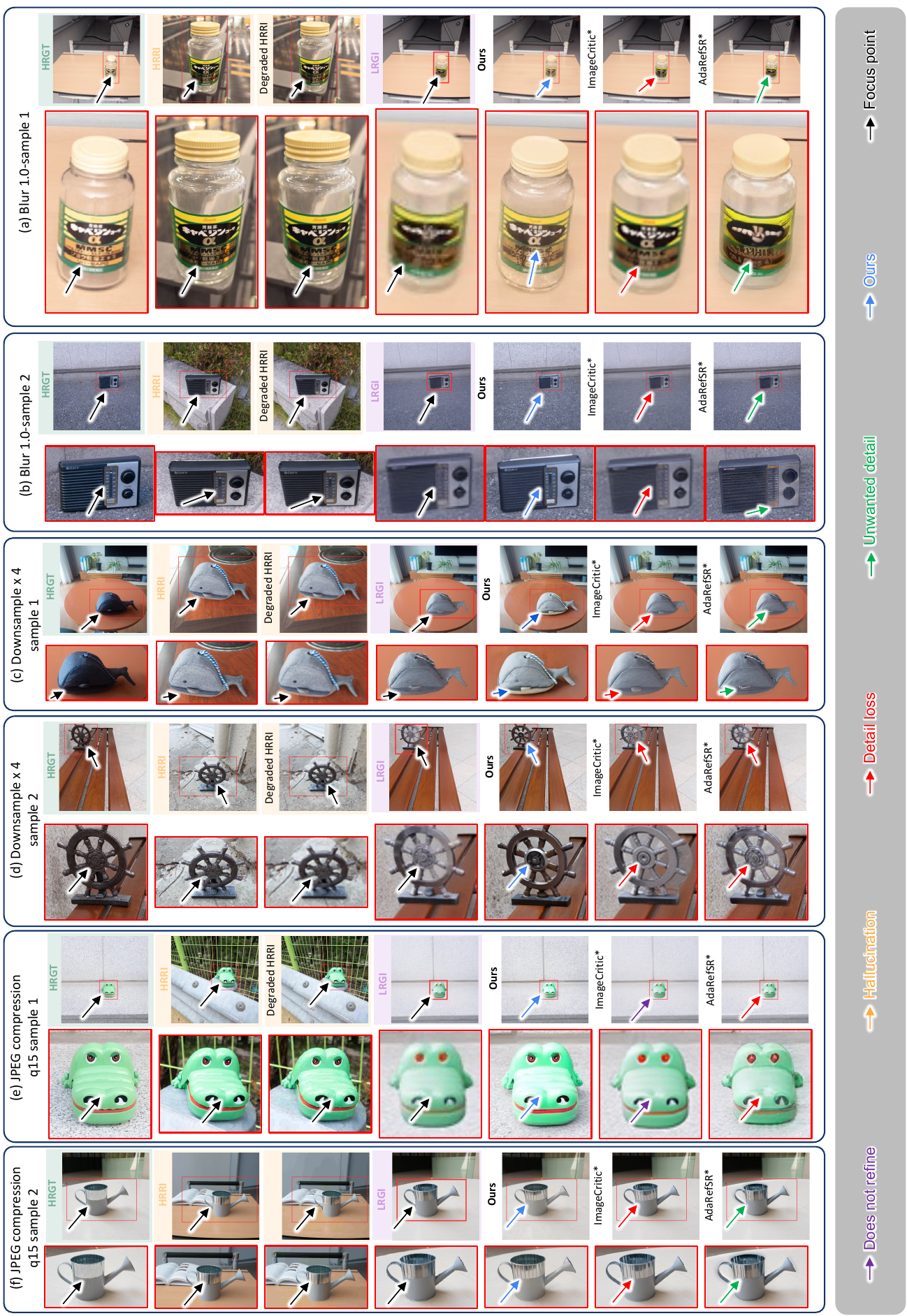}
  \caption{Qualitative comparison under degraded HRRI conditions.
  Rows (a) and (b) show Gaussian blur with $\sigma=1$, rows (c) and
  (d) show $4\times$ downsampling, and rows (e) and (f) show JPEG
  compression with $q=15$. Each degradation is applied only to the
  HRRI, while the LRGI remains unchanged. We compare RefGC-SR$^2$
  with ImageCritic~\citep{ouyang2025consistency} and
  AdaRefSR~\citep{wang2026trust}. The legend on the right explains
  the arrows. Best viewed zoomed in.}
  \label{fig:hrri_degradation}
\end{figure*}

\paragraph{Qualitative Results}
\jhzz{Fig.~\ref{fig:hrri_degradation} presents example outputs obtained with degraded HRRIs under the three evaluated conditions. These examples complement the quantitative comparisons with clean references by showing the resulting object appearance and fine details when degraded reference information is used.}

\begin{table}[h!]
    \centering
    \caption{
        \dah{Evaluation on 100 triplets from independent sources. HRRI--HRGT pairs are collected from sources excluded from the training set and original benchmark, while LRGIs are generated using the frozen DipRefGC model. An $*$ denotes a baseline fine-tuned on the RefGC-SR$^2$ Dataset. The best results are shown in bold.}
    }
    \label{tab:independent_sources}
    \small
    \setlength{\tabcolsep}{6pt}
    \begin{tabular}{lccccc}
        \toprule
        Method
        & CLIP-I$\uparrow$
        & DINO$\uparrow$
        & PSNR$\uparrow$
        & SSIM$\uparrow$
        & LPIPS$\downarrow$ \\
        \midrule
        DiT4SR$^*$
        & 0.8311 & 0.6456 & 14.32 & 0.4683 & 0.4850 \\
        TSD-SR$^*$
        & 0.8490 & 0.7334 & 16.30 & 0.4148 & 0.4835 \\
        ReFIR
        & 0.8715 & 0.7604 & 16.38 & 0.5561 & 0.3897 \\
        AdaRefSR$^*$
        & 0.8715 & 0.7675 & 17.07 & 0.5634 & 0.3366 \\
        OmniPaint
        & 0.8580 & 0.7006 & 14.24 & 0.5405 & 0.4659 \\
        ImageCritic$^*$
        & 0.8895 & 0.7871 & 16.19 & 0.5765 & 0.4022 \\
        RefGC-SR$^2$ (Ours)
        & \textbf{0.8911}
        & \textbf{0.8059}
        & \textbf{17.53}
        & \textbf{0.5956}
        & \textbf{0.3062} \\
        \bottomrule
    \end{tabular}
\end{table}

\subsection{Evaluation on Independent Sources and RefGC Outputs}
\label{subsec:independent_sources}

\dah{Our original benchmark contains non-overlapping HRRI--HRGT pairs with RefGC-SR$^2$ Dataset, but shares source datasets and the DipRefGC generator with training. We therefore evaluate two additional settings: independent reference pairs with DipRefGC, and independent reference pairs with off-the-shelf RefGC generators~\citep{huang2025dreamfuse, song2026insert, zhang2025freecus, feng2026personalize}.}

\paragraph{Independent Reference Pairs}
\dah{We evaluate 100 triplets constructed from HRRI--HRGT pairs collected from Wikimedia Commons~\citep{wikimedia}, ABO~\citep{collins2022abo}, DreamBench++~\citep{peng2024dreambench++}, and DreamBooth~\citep{ruiz2023dreambooth}. These sources are excluded from both our training set and original benchmark. We generate the corresponding LRGIs using the frozen DipRefGC model and follow the evaluation protocol of Table~\ref{tab:main_quan}. As shown in Table~\ref{tab:independent_sources}, RefGC-SR$^2$ achieves the best results across all five metrics. This evaluation supports generalization to pairs from unseen sources.}

\begin{table}[hbt]
    \centering
    \caption{\dah{Evaluation on 80 independent-source HRRI--HRGT pairs with off-the-shelf RefGC outputs. Each pair is processed by two generators per setting, yielding 160 compositing and 160 customization inputs. CLIP-I and DINO compare outputs with the HRRI using object bounding-box crops for compositing and full images for customization. Scores are averaged over the pooled outputs within each setting. An asterisk denotes a baseline fine-tuned on the RefGC-SR$^2$ Dataset. The best results are shown in bold.}}
\label{tab:independent_refgc}
    \label{tab:independent_refgc}
    \small
    \setlength{\tabcolsep}{8pt}
    \begin{tabular}{lcccc}
        \toprule
        & \multicolumn{2}{c}{Compositing}
        & \multicolumn{2}{c}{Customization} \\
        \cmidrule(lr){2-3}
        \cmidrule(lr){4-5}
        Method
        & CLIP-I$\uparrow$
        & DINO$\uparrow$
        & CLIP-I$\uparrow$
        & DINO$\uparrow$ \\
        \midrule
        DiT4SR$^*$
        & 0.8324 & 0.6653 & 0.8085 & 0.6173 \\
        TSD-SR$^*$
        & 0.8547 & 0.7483 & 0.8196 & 0.6543 \\
        ReFIR
        & 0.8700 & 0.7671 & 0.8359 & 0.6633 \\
        AdaRefSR$^*$
        & 0.8708 & 0.7634 & 0.8346 & 0.6617 \\
        OmniPaint
        & 0.8376 & 0.6836 & 0.8269 & 0.6604 \\
        ImageCritic$^*$
        & 0.8921 & 0.8024 & 0.8605 & 0.7140 \\
        RefGC-SR$^2$ (Ours)
        & \textbf{0.8928}
        & \textbf{0.8056}
        & \textbf{0.8610}
        & \textbf{0.7168} \\
        \bottomrule
    \end{tabular}
\end{table}
\paragraph{Independent Reference Pairs with Off-the-shelf RefGC Outputs}
\dah{For this evaluation, we select 80 HRRI--HRGT pairs from a 200-pair independent-source pool~\citep{wikimedia, collins2022abo, peng2024dreambench++, ruiz2023dreambooth, openverse}. We generate LRGIs using DreamFuse~\citep{huang2025dreamfuse} and InsertAnything~\citep{song2026insert} for compositing, and FreeCus~\citep{zhang2025freecus} and PersonalizeAnything~\citep{feng2026personalize} for customization. Each generator processes all 80 pairs, yielding 160 inputs per setting. Both the source datasets and input generators differ from those used to construct our training triplets.}

\dah{Because these generators do not enforce pose alignment with the HRGT, we report CLIP-I~\citep{radford2021learning} and DINO~\citep{oquab2023dinov2} similarity to the HRRI rather than pixel-aligned reconstruction metrics. For compositing, we use separate object bounding-box crops for the output and HRRI, with a 10\% margin on each side. The output crop is defined by the HRGT object mask, while the reference crop is defined by the HRRI mask; background pixels within the crops are retained. For customization, we compare the full output and HRRI images because the generated layout may differ from the target. Within each setting, we report the mean of the per-image scores over all 160 outputs.}

\dah{Table~\ref{tab:independent_refgc} shows that RefGC-SR$^2$ achieves the highest CLIP-I and DINO scores in both settings. The margins over ImageCritic$^*$~\citep{ouyang2025consistency}, which shares the same backbone~\citep{labs2025flux} with ours, are modest: 0.0007 CLIP-I and 0.0032 DINO for compositing, and 0.0005 CLIP-I and 0.0028 DINO for customization. These results demonstrate that RefGC-SR$^2$ generalizes well
beyond the training distribution, including both unseen reference sources and inputs generated by off-the-shelf RefGC models.}

\section{Multi-Object Evaluation}
\label{supsec:multi_object}

\paragraph{Benchmark Construction}
\jhz{To evaluate generalization beyond single-object scenes, we construct a multi-object benchmark using EntitySeg~\citep{qi2022fine}. We select 36 real images containing three to five distinct objects and use each original image as the HRGT. For each HRGT, we generate an HRRI using Gemini 3.1 Flash Image~\citep{google2026gemini31flashimage} by varying the viewpoint and scene context while preserving the identities of all objects. We manually inspect the generated pairs and exclude cases in which the HRRI deviates substantially from the HRGT or fails to preserve the object identities. The corresponding LRGI is synthesized using DipRefGC following the same procedure used for our main dataset.}

\paragraph{Evaluation Protocol}
\jhz{All models are evaluated without retraining or finetuning on multi-object data, following the same evaluation protocol as Table~\ref{tab:main_quan}.}

\begin{figure*}[p]
  \centering
  \includegraphics[width=\linewidth]{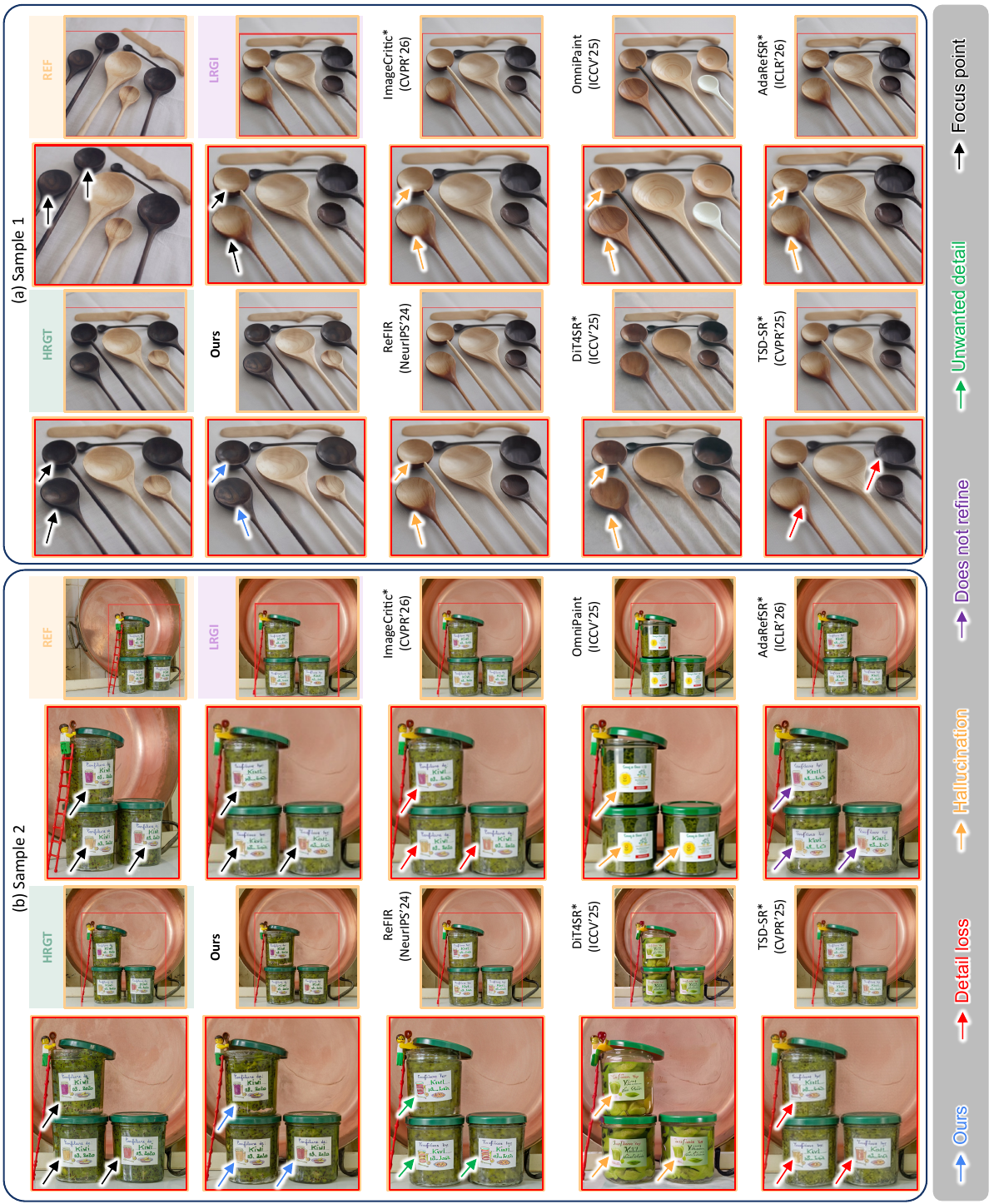}
  \caption{Qualitative comparison in the multi-object setting. We
  compare our RefGC-SR\textsuperscript{2} against
  ImageCritic~\citep{ouyang2025consistency},
  OmniPaint~\citep{yu2025omnipaint}, AdaRefSR~\citep{wang2026trust},
  ReFIR~\citep{guo2024refir}, TSD-SR~\citep{dong2025tsd}, and
  DiT4SR~\citep{duan2025dit4sr}. Please refer to the legend below the
  figure for the meaning of each arrow. Best viewed zoomed in.}
  \label{fig:multi-object}
\end{figure*}

\begin{table*}[t]
\centering
\caption{Quantitative comparison on the multi-object benchmark with 36 scenes containing three to five objects each. All methods are evaluated without multi-object training. The best and second-best results are shown in \textbf{bold} and \underline{underline}, respectively. An asterisk ($*$) indicates a baseline finetuned on the RefGC-SR$^2$ Dataset.}
\label{tab:multi_object}

\begingroup
\small
\setlength{\tabcolsep}{6pt}
\renewcommand{\arraystretch}{0.95}

\resizebox{0.60\textwidth}{!}{%
\begin{tabular}{lccccc}
\toprule
Model
& CLIP-I
& DINO
& PSNR
& SSIM
& LPIPS \\
\midrule
DiT4SR$^*$      
& 0.8920 & 0.7668 & 16.0412 & 0.5940 & 0.3617 \\

TSD-SR$^*$      
& 0.8996 & 0.8031 & 16.7320 & 0.5759 & 0.3876 \\

ReFIR           
& \underline{0.9280}
& \underline{0.8483}
& 15.7616
& 0.5494
& 0.3422 \\

AdaRefSR$^*$    
& 0.9234
& 0.8336
& \underline{16.9844}
& \underline{0.5992}
& \underline{0.2889} \\

OmniPaint       
& 0.8905 & 0.7722 & 14.6862 & 0.5493 & 0.5020 \\

ImageCritic$^*$ 
& 0.9188 & 0.8424 & 16.2762 & 0.5931 & 0.4158 \\
\midrule
RefGC-SR$^2$ (Ours)
& \textbf{0.9378}
& \textbf{0.8720}
& \textbf{18.6452}
& \textbf{0.6416}
& \textbf{0.2194} \\
\bottomrule
\end{tabular}%
}

\endgroup
\end{table*}

\paragraph{Quantitative Result}
\jhz{As shown in Table~\ref{tab:multi_object}, the RefGC-SR$^2$ model achieves state-of-the-art performance across all five metrics, demonstrating strong generalization to scenes containing multiple object instances without multi-object training.}

\paragraph{Qualitative Results}
\jhz{Fig.~\ref{fig:multi-object} provides additional qualitative comparisons under the multi-object setting. 
% As shown in Fig.~\ref{fig:multi-object}(a), when refining a scene containing multiple spoons, the competing methods fail to recover the black appearance of the two rightmost spoons, whereas our method successfully restores their reference-consistent appearance. 
As shown in Fig.~\ref{fig:multi-object}(a), when refining a scene containing multiple spoons, the competing methods fail to recover the black appearance of the two leftmost spoons, whereas our method successfully restores their reference-consistent appearance.
In Fig.~\ref{fig:multi-object}(b), the competing methods introduce hallucinated content and unwanted details. In contrast, the proposed RefGC-SR$^2$ restores the multiple objects without introducing these artifacts.}

\section{Qualitative Results on Outputs from Commercial and Open-Source Models}
\label{sup:commercial_comp}

% In this section, we provide qualitative comparisons to demonstrate the performance of our RefGC-SR\textsuperscript{2} model. We consider two commercial (proprietary) models, Gemini~2.5 Flash Image~\citep{google2025geminiflashimage} and GPT-Image~1.5~\citep{openai2025gptimage}, and one open-source model, Qwen-Image-Edit~\citep{wu2025qwenimage}. We first collect a reference image (HRRI). We then employ the Qwen3-VL-Instruct model~\citep{bai2025qwen3} to generate text prompts describing a subject using or wearing a specific object. Using the resulting HRRI and text prompts as inputs to the two commercial models (Gemini~2.5 Flash Image~\citep{google2025geminiflashimage} and GPT-Image~1.5~\citep{openai2025gptimage}) and the open-source model (Qwen-Image-Edit~\citep{wu2025qwenimage}), we generate images and treat the resulting images as LRGIs. Given these LRGIs and HRRIs, we qualitatively compare the results of our RefGC-SR\textsuperscript{2} model against six competing models: ImageCritic~\citep{ouyang2025consistency}, OmniPaint~\citep{yu2025omnipaint}, AdaRefSR~\citep{wang2026trust}, ReFIR~\citep{guo2024refir}, TSD-SR~\citep{dong2025tsd}, and DiT4SR~\citep{duan2025dit4sr}.

\begin{figure*}[p]
  \centering
  \includegraphics[width=\linewidth]{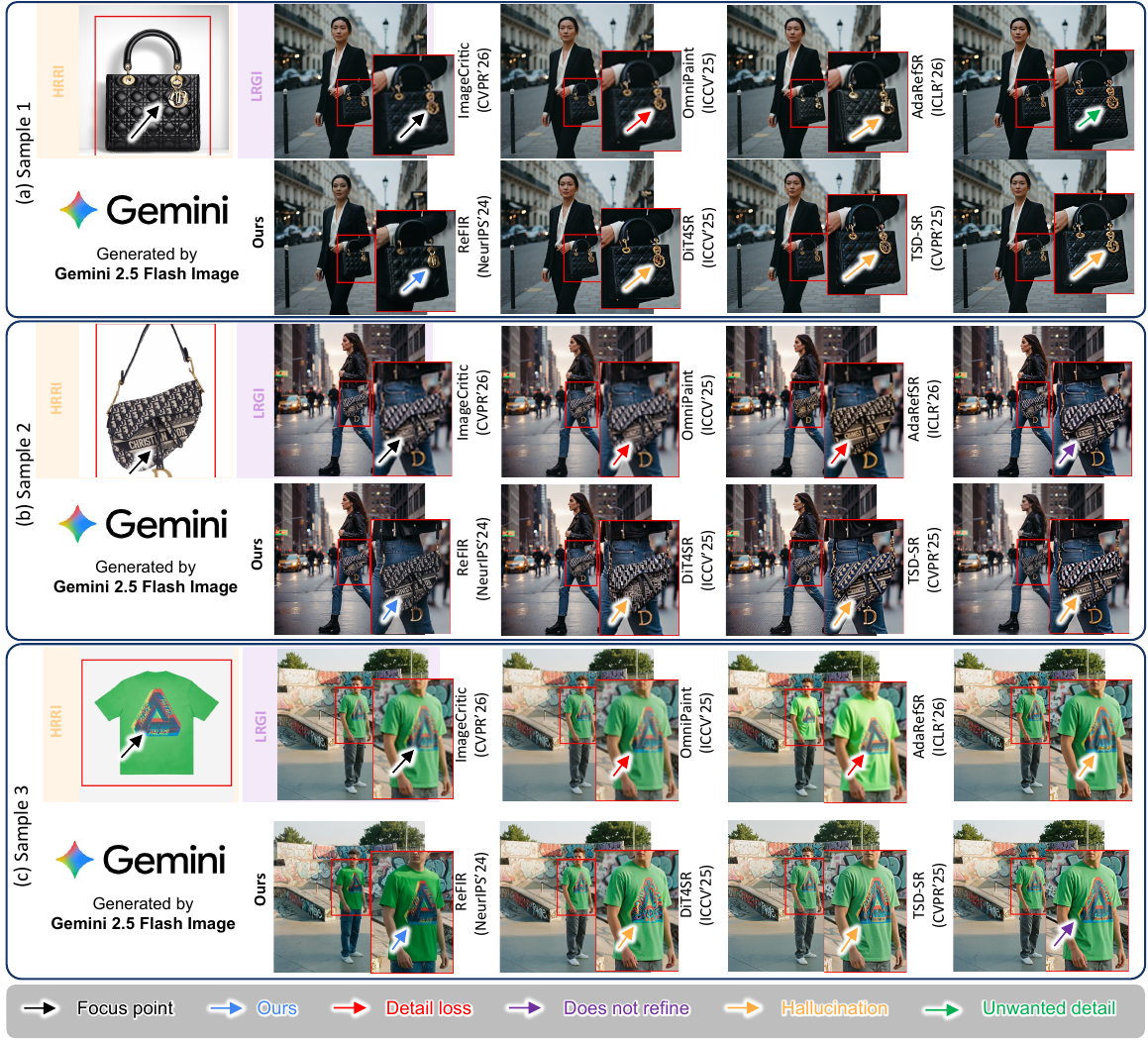}
  \caption{Qualitative comparison on LRGIs generated by the commercial
  model Gemini~2.5 Flash Image~\citep{google2025geminiflashimage}. We
  compare our RefGC-SR\textsuperscript{2} against
  ImageCritic~\citep{ouyang2025consistency},
  OmniPaint~\citep{yu2025omnipaint}, AdaRefSR~\citep{wang2026trust},
  ReFIR~\citep{guo2024refir}, TSD-SR~\citep{dong2025tsd}, and
  DiT4SR~\citep{duan2025dit4sr}. Please refer to the legend below the
  figure for the meaning of each arrow. Best viewed zoomed in.}
  \label{fig:comp_gemini}
\end{figure*}

\begin{figure*}[p]
  \centering
  \includegraphics[width=\linewidth]{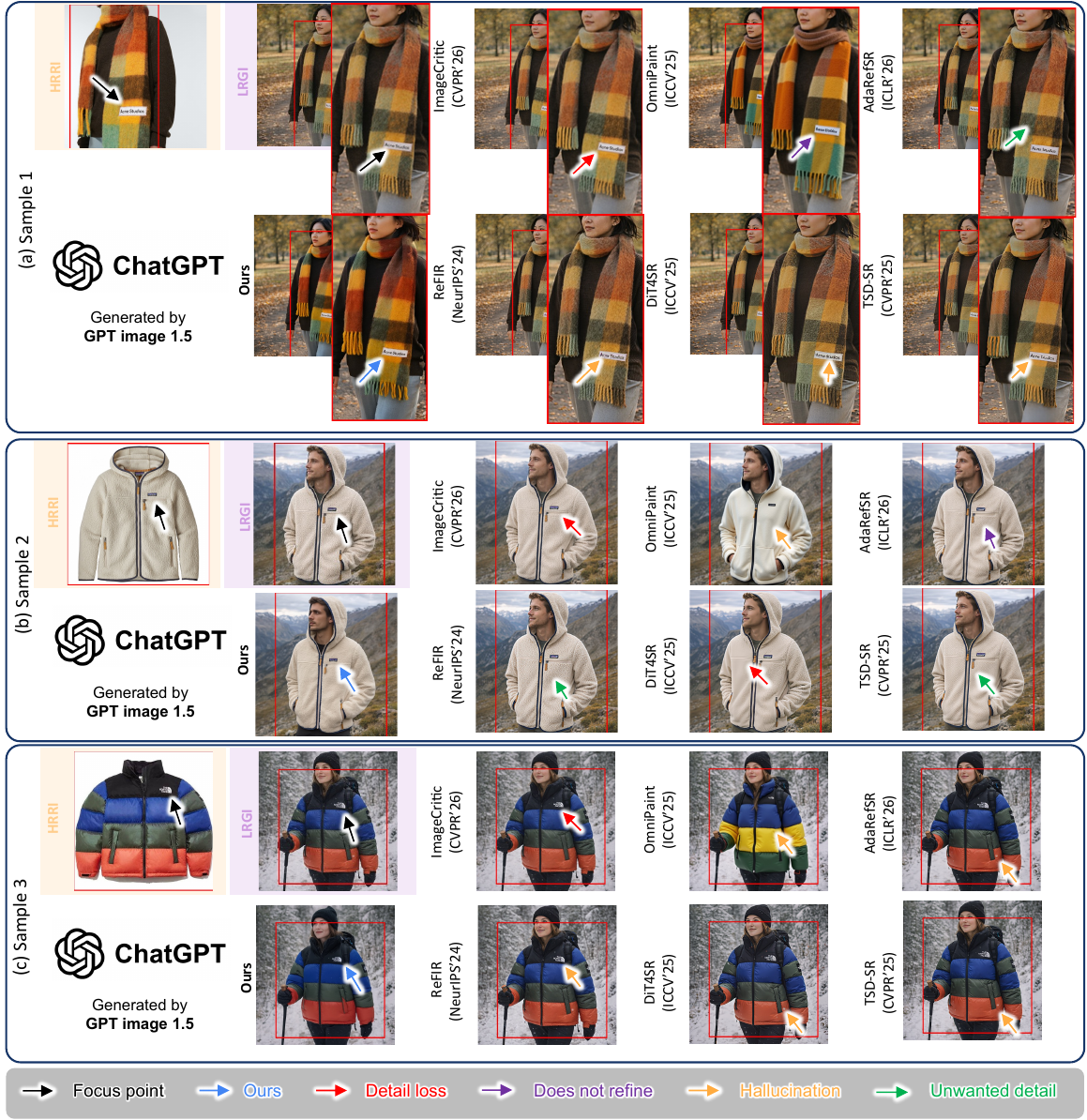}
  \caption{Qualitative comparison on LRGIs generated by the commercial
  model GPT-Image~1.5~\citep{openai2025gptimage}. We compare our
  RefGC-SR\textsuperscript{2} against
  ImageCritic~\citep{ouyang2025consistency},
  OmniPaint~\citep{yu2025omnipaint}, AdaRefSR~\citep{wang2026trust},
  ReFIR~\citep{guo2024refir}, TSD-SR~\citep{dong2025tsd}, and
  DiT4SR~\citep{duan2025dit4sr}. Please refer to the legend below the
  figure for the meaning of each arrow. Best viewed zoomed in.}
  \label{fig:comp_gpt}
\end{figure*}

\begin{figure*}[p]
  \centering
  \includegraphics[width=\linewidth]{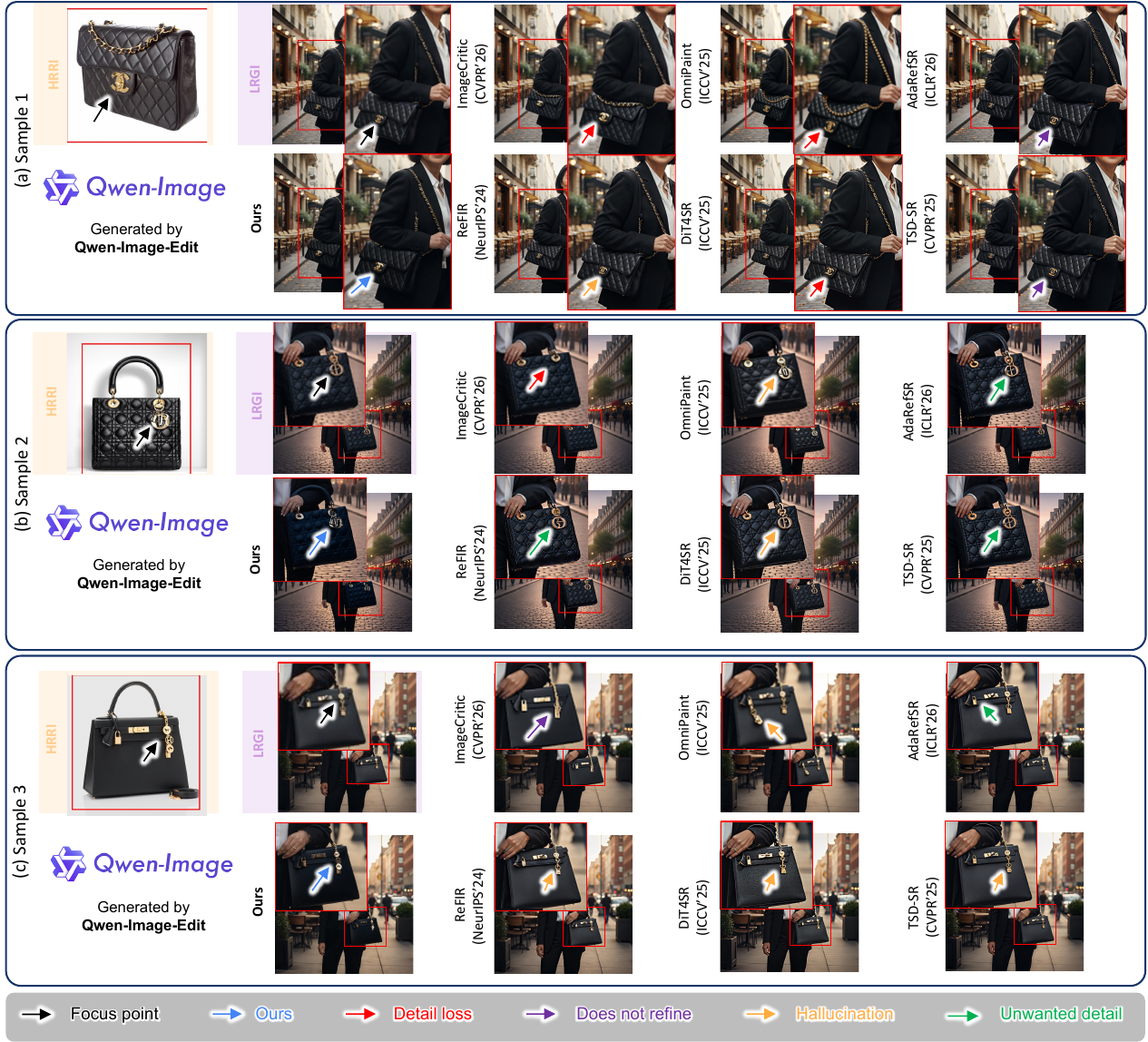}
  \caption{Qualitative comparison on LRGIs generated by the open-source
  model Qwen-Image-Edit~\citep{wu2025qwen}. We compare our
  RefGC-SR\textsuperscript{2} against
  ImageCritic~\citep{ouyang2025consistency},
  OmniPaint~\citep{yu2025omnipaint}, AdaRefSR~\citep{wang2026trust},
  ReFIR~\citep{guo2024refir}, TSD-SR~\citep{dong2025tsd}, and
  DiT4SR~\citep{duan2025dit4sr}. Please refer to the legend below the
  figure for the meaning of each arrow. Best viewed zoomed in.}
  \label{fig:comp_qwen}
\end{figure*}

Fig.~\ref{fig:comp_gemini} provides a qualitative comparison across seven models, including our RefGC-SR\textsuperscript{2} Model, when the outputs of Gemini~2.5 Flash Image are treated as LRGIs. Similarly, Fig.~\ref{fig:comp_gpt} reports the qualitative comparison when the outputs of GPT-Image~1.5 are treated as LRGIs, and Fig.~\ref{fig:comp_qwen} presents the qualitative comparison when the outputs of Qwen-Image-Edit are treated as LRGIs. 
As shown in Figs.~\ref{fig:comp_gemini} - \ref{fig:comp_qwen}, our proposed RefGC-SR$^2$ faithfully restores fine details and enhances resolution while simultaneously removing generative artifacts, whereas the competing models commonly suffer from either hallucinating undesired details or failing to recover fine details.
These results demonstrate that our proposed RefGC-SR\textsuperscript{2} Model generalizes well even to image pairs drawn from distributions different from those seen during training.

% ===== Gemini 2.5 Flash Image qualitative comparison =====
\clearpage
% ===== GPT-Image 1.5 qualitative comparison =====
\clearpage
% ===== Qwen-Image-Edit qualitative comparison =====
\clearpage

\section{User Study Protocols and Analysis}
\label{supsec:user_study}

\begin{figure*}[h!]
	\centering
	\includegraphics[width=1.0\linewidth]{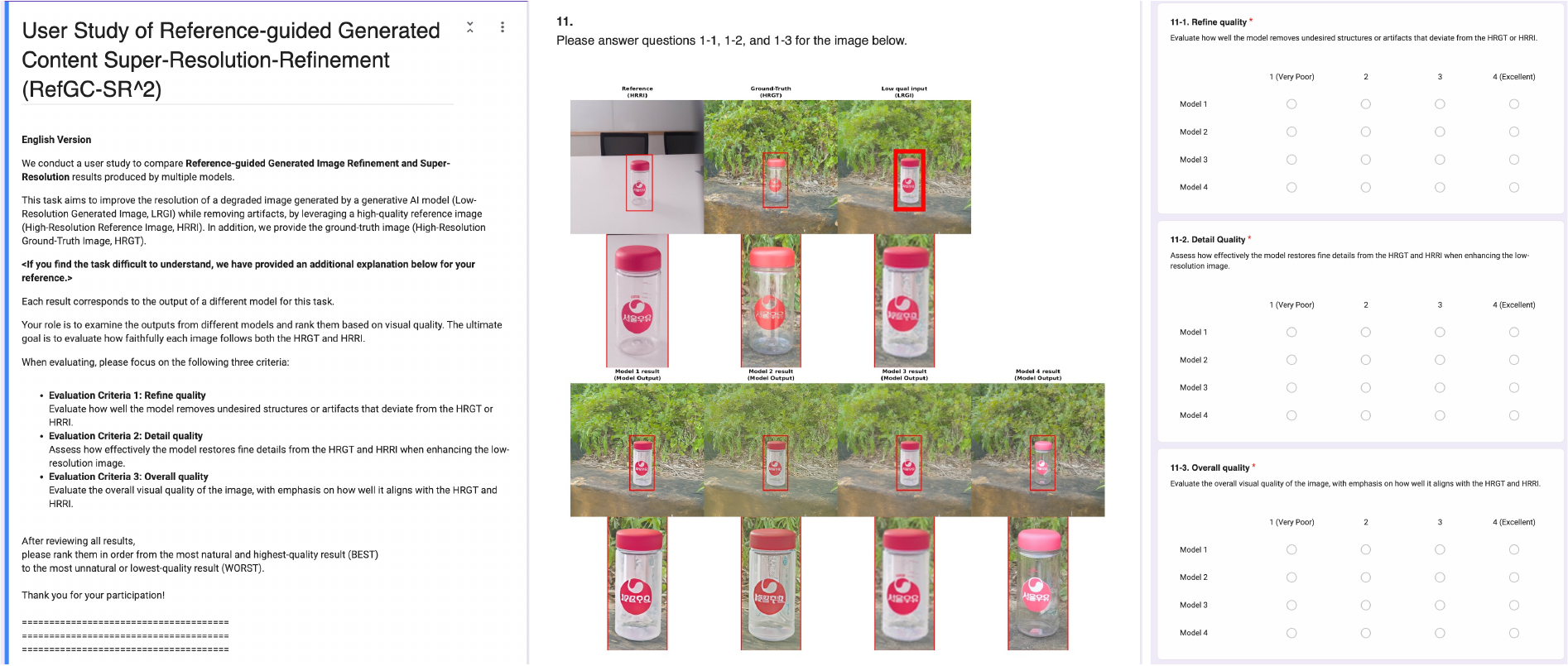}
	\caption{User study questionnaire interface (Google Forms). Left: task instructions and three evaluation criteria. Center: a sample question showing HRRI, HRGT, LRGI, and four anonymized model outputs (``Model 1''--``Model 4'') with zoomed-in object crops. Right: 4-point Likert scale rating for refine/detail/overall quality.}
    \label{fig:user_study_form}
\end{figure*}

\begin{figure*}[h!]
	\centering
	\includegraphics[width=1.0\linewidth]{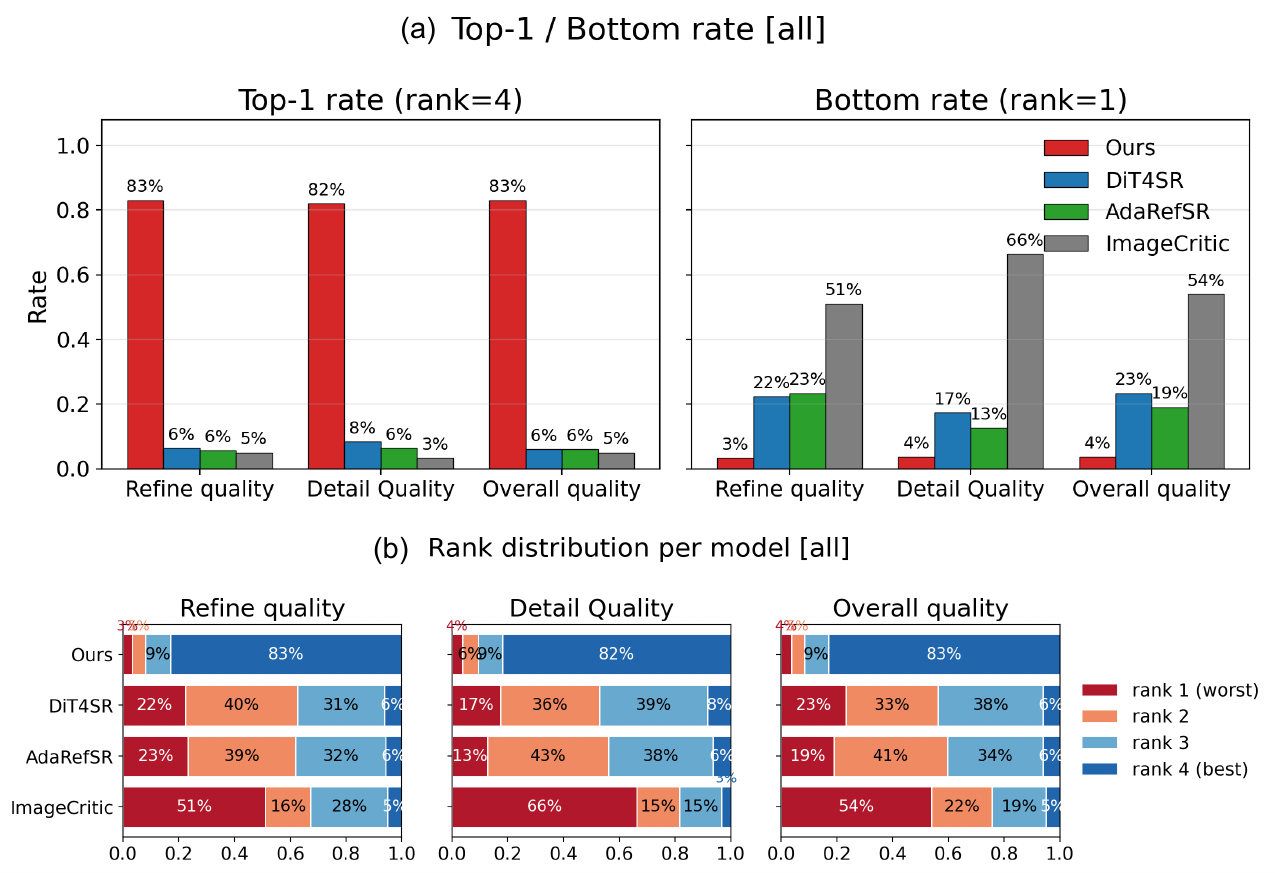}
	\caption{Detailed user study results on the In-the-wild Benchmark. (a) Top-1 (rank=4) and bottom (rank=1) rates per model across three evaluation criteria. (b) Full rank distribution per model. Our model dominates the best rank (82--83\%) and is rarely ranked worst (3--4\%), while ImageCritic incurs the highest worst-rank rates (51\%/66\%/54\%) due to its oversmoothing tendency.}
    \label{fig:user_study_supple}
\end{figure*}

\subsection{Protocol}
\label{sec:user_study_protocol}

We provide additional details on the user study reported in Sec.~\ref{sec:experiment}-\textbf{User Study}. The study was conducted as a blind, side-by-side comparison via a web-based questionnaire (Google Forms) and took approximately 20 minutes per participant. Participation was voluntary, with no monetary compensation; participants were informed about the purpose and approximate duration of the study before consenting to participate, and could withdraw at any time. No personally identifiable information was collected.

\noindent\textbf{Question Composition} The questionnaire consisted of 20 questions, each corresponding to one in-the-wild test case. As shown in Fig.~\ref{fig:user_study_form}, each question presents three reference images at the top (HRRI, HRGT, and LRGI) followed by four model outputs anonymized as ``Model 1''--``Model 4''. The four outputs correspond to one method from each of SR (DiT4SR~\citep{duan2025dit4sr}), RefSR (AdaRefSR~\citep{wang2026trust}), RefGCR (ImageCritic~\citep{ouyang2025consistency}), and our RefGC-SR$^2$ Model. To facilitate fine-grained comparison, every image is accompanied by a zoomed-in crop of the object region, with the relevant region highlighted by a red bounding box.

\noindent\textbf{Evaluation Criteria} For each question, participants rate all four model outputs on three independent criteria using a 4-point Likert scale (1: Very Poor, 2, 3, 4: Excellent): (i) \textit{Refine quality}: how well the model removes undesired structures or artifacts that deviate from the HRGT or HRRI; (ii) \textit{Detail quality}: how effectively the model restores fine details from the HRGT and HRRI when enhancing the low-resolution image; (iii) \textit{Overall quality}: the overall visual quality of the image, with emphasis on alignment with the HRGT and HRRI. The full text of the instructions provided to participants is shown on the left panel of Fig.~\ref{fig:user_study_form}.

\noindent\textbf{Scoring} From the collected ratings, we compute two complementary statistics: (i) \textit{top-1 rate}, defined as the proportion of (participant, question) pairs in which the model received the highest score (rank 4) among the four candidates, and (ii) \textit{bottom rate}, defined analogously for rank 1. The aggregated results across 16 participants $\times$ 20 questions are reported in the main paper (Fig.~\ref{fig:user_study}) and in further detail in Fig.~\ref{fig:user_study_supple}.

\subsection{Detailed Analysis}
\label{sec:user_study_analysis}

Fig.~\ref{fig:user_study_supple} provides two complementary views of the results: (a) top-1 and bottom rates per model under each criterion, and (b) the full rank distribution per model.

\noindent\textbf{Overall Preference} As reported in the main paper, our model achieves a top-1 rate of 82--83\% across all three criteria, while baselines remain below 9\% (Fig.~\ref{fig:user_study_supple}-(a), left). Symmetrically, our model is rarely ranked worst (3--4\%), in contrast to baselines that incur substantially higher worst-rank rates (Fig.~\ref{fig:user_study_supple}-(a), right). The full rank distribution in Fig.~\ref{fig:user_study_supple}-(b) further confirms that our model dominates rank 4 (best) across all criteria, while DiT4SR and AdaRefSR are predominantly distributed across ranks 2 and 3, indicating mid-tier performance.

\noindent\textbf{Discrepancy with Quantitative Metrics for ImageCritic} A particularly notable observation is the contrast between ImageCritic's quantitative and human evaluation results. While ImageCritic ranks second-best on the RefGC-SR$^2$ Benchmark in Table~\ref{tab:main_quan}, Fig.~\ref{fig:user_study_supple}-(b) shows that it receives the worst rank in 51\%, 66\%, and 54\% of cases for refine, detail, and overall quality, respectively, the worst among all compared models. We attribute this discrepancy to ImageCritic's tendency to produce \textit{oversmoothed outputs}: while smoothing tends to align well with HRGT in pixel-level metrics (PSNR, SSIM) and feature-level similarity (CLIP-I, DINO), human evaluators perceive the resulting loss of fine-grained texture and high-frequency detail as a clear quality degradation. This finding underscores the importance of complementing quantitative metrics with human evaluation for tasks involving fine detail recovery, and supports the design of our frequency-aware FreqMoLE module, which is explicitly aimed at preserving high-frequency information from the HRRI.

\begin{figure*}[h!]
	\centering
	\includegraphics[width=1.0\linewidth]{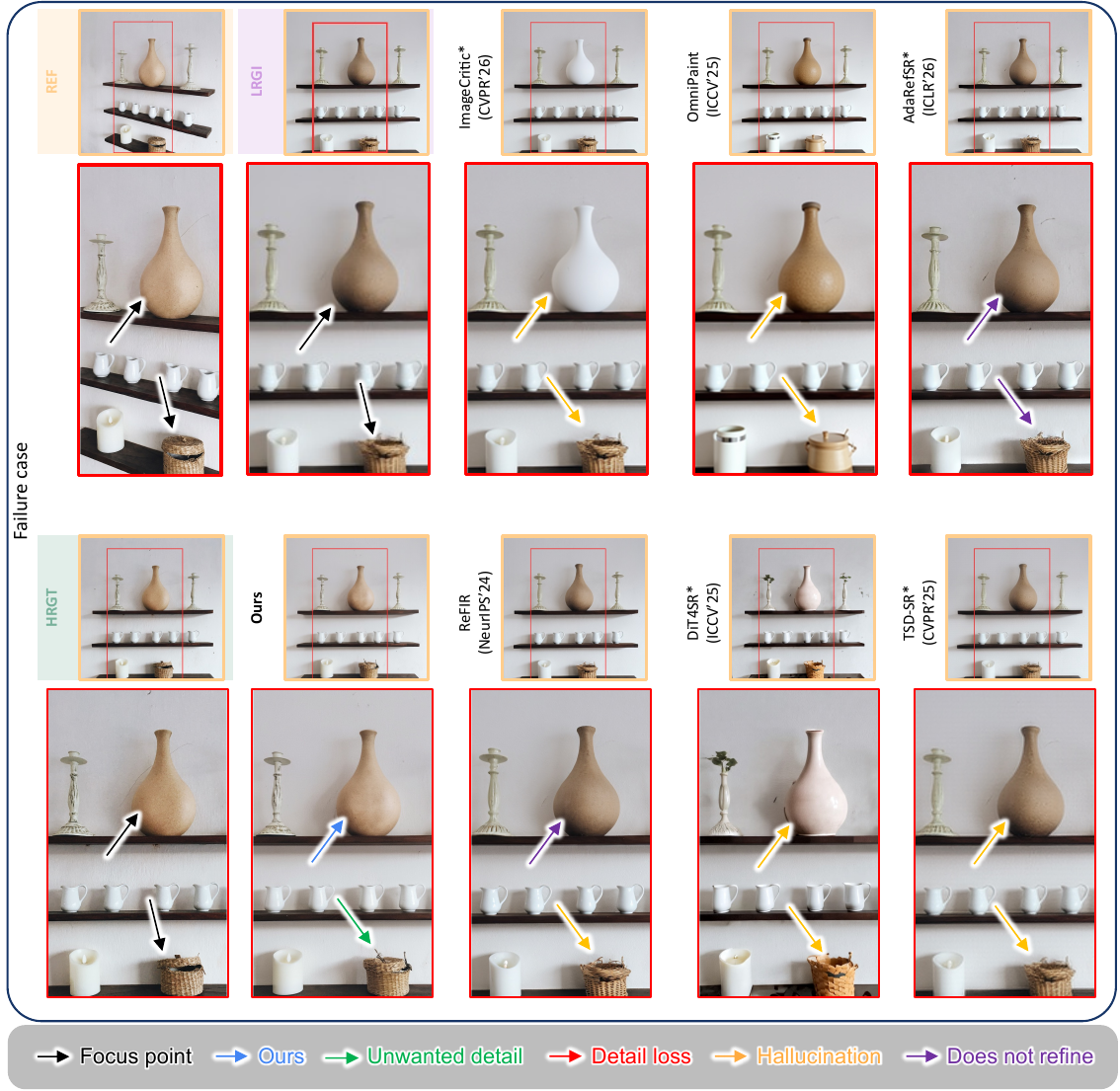}
	\caption{Failure case in a multi-object scene. Our RefGC-SR$^2$ model restores the reference-consistent appearance of the vase on the upper shelf but still introduces unwanted details around the basket on the lower shelf, illustrating its limitations in multi-object refinement.}
    \label{fig:multi-obj-failure}
\end{figure*}

\section{Limitations}
\label{sec:limitations}

% \jhz{Our current formulation is specialized for $4\times$ upscaling. Due to GPU memory and computational budget constraints, we were unable to train and evaluate the model across multiple upscaling factors. Extending the proposed framework to arbitrary-scale super-resolution is therefore an important direction for future work. In addition, although our model demonstrates generalization on the multi-object benchmark (Appendix~\ref{supsec:multi_object}), it was trained only on object-centric single-object data. Its performance may consequently remain limited in more complex multi-object scenes involving severe occlusions or interactions between objects. Future work will incorporate multi-object training data to further improve robustness in these settings.}
\jhzz{Although trained only on object-centric single-object data, RefGC-SR$^2$ achieves the best performance across all five metrics on the multi-object benchmark (Appendix~\ref{supsec:multi_object}, Table~\ref{tab:multi_object}). Nevertheless, its outputs are not always free of artifacts. In Fig.~\ref{fig:multi-obj-failure}, competing methods introduce errors in the clay vase on the upper shelf, the basket on the lower shelf, or both. Our model restores the vase's reference-consistent appearance but still produces unwanted details around the basket. Thus, while RefGC-SR$^2$ performs better overall, it can still produce local artifacts in multi-object scenes. 
We leave the construction of a multi-object dataset and subsequent training of RefGC-SR$^2$ on that dataset to future work.
}

\newpage

\end{document}